\documentclass[10pt,twocolumn,letterpaper]{article}

\usepackage[pagenumbers]{iccv} 

\definecolor{iccvblue}{rgb}{0.21,0.49,0.74}
\usepackage[pagebackref,breaklinks,colorlinks,allcolors=iccvblue]{hyperref}
\usepackage{multirow}
\usepackage{multicol}
\usepackage{xcolor}

\title{ Multi-modal Multi-platform Person Re-Identification: Benchmark and Method}

\author{Ruiyang Ha$^{1,\dagger}$\quad  Songyi Jiang$^{1,\dagger}$ \quad Bin Li$^1$\quad Bikang Pan$^1$\quad Yihang Zhu$^1$\quad \\Junjie Zhang$^2$\quad Xiatian Zhu$^3$\quad Shaogang Gong$^4$ Jingya Wang$^{1,}$\thanks{Corresponding author. } \\
$^1$ShanghaiTech University,
$^2$The Xi'an Jiaotong-Liverpool University\\
$^3$University of Surrey
$^4$Queen Mary University London\\
}

\begin{document}
\twocolumn[{ \renewcommand\twocolumn[1][]{#1}
    \maketitle
    \begin{center}
        \centering \includegraphics[width=\linewidth]{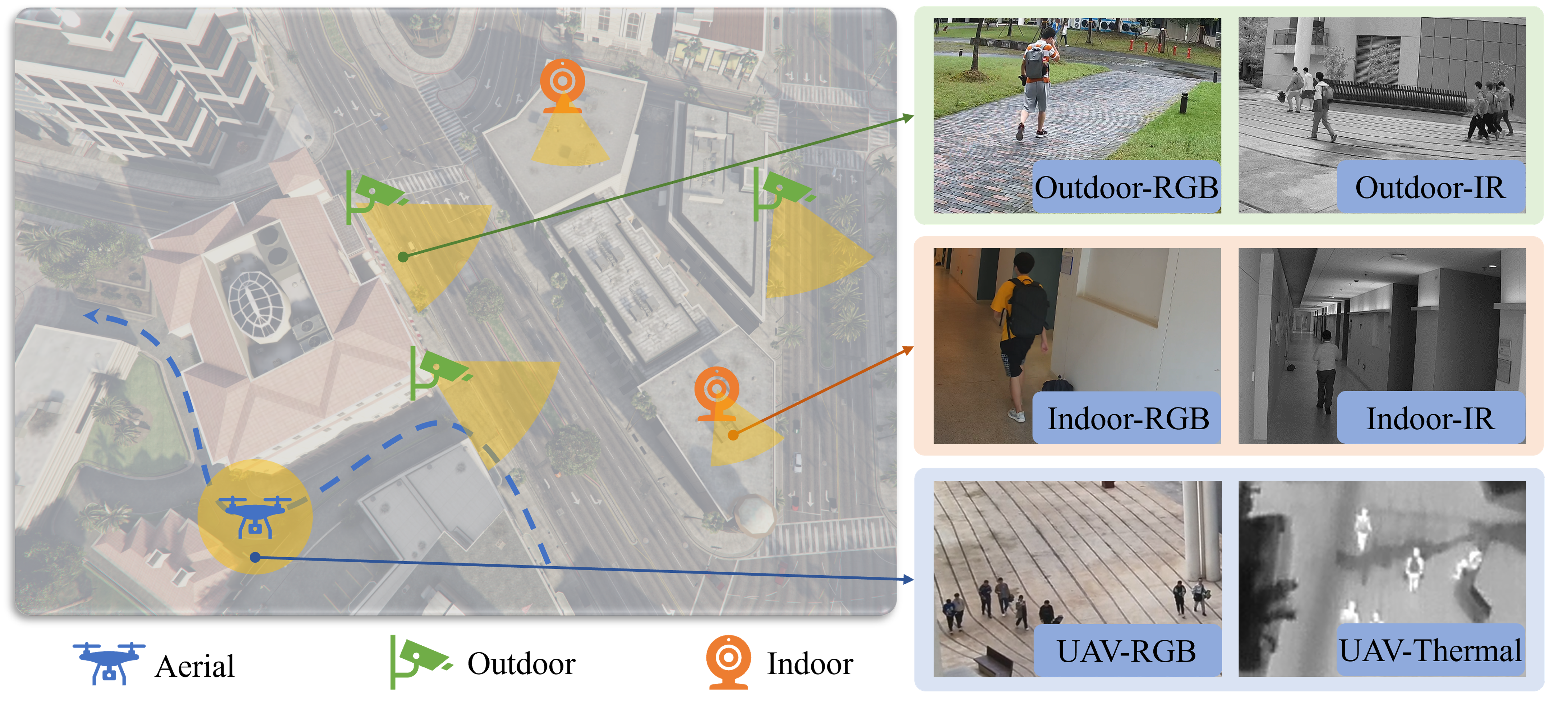}
        \captionof{figure}{A new MP-ReID dataset is represented by a conceptual diagram, showcasing the inclusion of six ground RGB cameras, six ground infrared cameras, one UAV RGB camera and one UAV thermal camera. The data is collected from a variety of environments, including outdoor and indoor settings, as well as from UAV cameras for aerial perspectives. This integration of diverse data sources and modalities creates a comprehensive and versatile dataset, aimed at enhancing research in multi-modal human perception.}
        \label{fig:examples in our MP-ReID}
    \end{center}
}]

\def\thefootnote{$\dagger$}\footnotetext{Equal contributions.}
\def\thefootnote{*}\footnotetext{Corresponding author.}

\begin{abstract}
Conventional person re-identification (ReID) research is often limited to single-modality sensor data from static cameras, which fails to address the complexities of real-world scenarios where multi-modal signals are increasingly prevalent. For instance, consider an urban ReID system integrating stationary RGB cameras, nighttime infrared sensors, and UAVs equipped with dynamic tracking capabilities. Such systems face significant challenges due to variations in camera perspectives, lighting conditions, and sensor modalities, hindering effective person ReID.
To address these challenges, we introduce the MP-ReID benchmark, a novel dataset designed specifically for multi-modality and multi-platform ReID. This benchmark uniquely compiles data from 1,930 identities across diverse modalities, including RGB, infrared, and thermal imaging, captured by both UAVs and ground-based cameras in indoor and outdoor environments.
Building on this benchmark, we introduce Uni-Prompt ReID, a framework with specific-designed prompts, tailored for cross-modality and cross-platform scenarios. Our method consistently outperforms state-of-the-art approaches, establishing a robust foundation for future research in complex and dynamic ReID environments. 
Our dataset are available at:
\url{https://mp-reid.github.io/}.
\end{abstract}
    
\section{Introduction}
    
    

Person Re-Identification (ReID)\cite{wu2019deep,ye2021deep,zheng2016person}  aims to identify and match pedestrians captured by non-overlapping, wide-field cameras. Traditional ReID approaches \cite{bai2020deep,chen2017beyond,li2017learning,liu2019deep,wei2018person} have primarily relied on static cameras and single-modality data, which limits their ability to handle the dynamic and diverse conditions of real-world environments. However, recent advances in diverse platforms and sensors, including vehicle-mounted cameras\cite{chen2021vehicle}, unmanned aerial vehicles (UAVs)\cite{zhang2020person,organisciak2021uav}, and wearable cameras\cite{nguyen2023aerial}, offer new opportunities to exploit multi-modality and multi-platform data. 
These developments enable data collection from various perspectives, environments, and lighting conditions, significantly enriching the data available for ReID systems and enhancing their adaptability in complex scenarios.
For instance, consider a 24/7 urban pedestrian ReID system integrating RGB cameras, nighttime infrared sensors, and UAVs with active tracking capabilities. Such a multi-modality, multi-platform framework provides comprehensive coverage, capturing data continuously for uninterrupted monitoring. UAVs, in particular, offer dynamic tracking and flexible viewpoints, while diverse modalities enhance situational awareness. By efficiently fusing data from different platforms and sensors, such systems can achieve a holistic understanding of the monitored environment, improving security and data collection efficiency.

Existing cross-modality datasets\cite{wu2017rgb, lin2022learning, zhang2023diverse, nguyen2017person} are limited to RGB and infrared modalities, collected primarily using static ground cameras, failing to address the complexities of real-world applications. 
Although UAV-based datasets like AG-ReID\cite{nguyen2023aerial} and AG-ReID v2\cite{nguyen2024ag} have been introduced, they focus only on aerial and ground RGB data and lack modality diversity. The absence of datasets combining multi-modality with multi-platform capabilities highlights a critical gap in the field.


To address this gap, we introduce the Multi-modality Multi-platform ReID (MP-ReID) benchmark, a novel dataset tailored for complex urban environments. As illustrated in Fig.~\ref{fig:examples in our MP-ReID}, MP-ReID comprises data collected from ground RGB cameras, ground infrared cameras, and UAVs equipped with RGB and thermal cameras, deployed across diverse indoor and outdoor locations. The dataset includes 1,930 identities and 136,156 manually annotated bounding boxes, ensuring robust support for research. Although the dataset size is constrained by the high costs of multi-modality and multi-platform data collection and annotation, MP-ReID surpasses existing cross-modality and cross-platform datasets in scale and diversity (Table~\ref{tab:freq}).

However, simultaneously addressing the challenges posed by modality discrepancies and ground-aerial disparities remains an extremely difficult problem. Existing methods primarily focus on bridging the modality gaps between infrared/RGB or thermal/RGB\cite{ye2021channel, ye2023channel, zhang2023diverse,fang2023visible,wang2022optimal,zhang2022modality,wu2024enhancing,pang2024inter,ge2025spectral,yin2025robust,jiang2025laboratory,yuan2025poses,niu2025chatreid,joseph2025clothes}, largely ignoring the additional complexities introduced by diverse capture devices. These complexities include variations in viewpoints, resolutions, and indoor/outdoor environments.
Leveraging the advancements in visual foundation models, we incorporate pretrained vision-language models such as CLIP and apply prompt learning to better handle complex scenarios involving multiple modalities and platforms. To this end, we propose Uni-Prompt ReID, a unified prompt learning framework that integrates modality-aware prompts and platform-aware prompts with visual-enhanced context into a unified text prompt structure. By doing so, our model achieves enhanced adaptability to diverse conditions, leading to significant improvements in pedestrian ReID performance across a wide range of challenging environments.
In summary, the main contributions of this work are as follows:

\begin{enumerate}

    

    \item We introduce MP-ReID, the first multi-modality multi-platform benchmark for person re-identification. MP-ReID integrates data from 14 cameras, including 6 ground-based RGB cameras, 6 ground-based infrared cameras, and a UAV equipped with both RGB and thermal sensors, simulating complex urban environments. This innovative setup enables comprehensive person ReID with unprecedented coverage, significantly improving recognition accuracy across diverse scenarios through the synergistic use of multiple modalities and platforms.
    \item To address the inherent challenges of multi-modality and multi-platform ReID tasks, we introduce Uni-Prompt ReID, a unified prompt learning framework. Our approach seamlessly integrates modality-aware and platform-aware prompts with visual-enhanced context into a unified text prompt structure. By leveraging these tailored components, Uni-Prompt ReID effectively bridges the gaps across diverse modalities and platforms, setting a new standard for robustness and adaptability in ReID tasks.
    
    \item We conducted extensive experiments on the newly introduced benchmark, demonstrating the superior performance of our proposed framework. To foster further research and ensure reproducibility, we will publicly release the MP-ReID dataset and provide all code for open access.
\end{enumerate}

\section{Related Work}
\label{sec:rel}

\subsection{Person Re-identification Dataset}

In the field of ReID, numerous datasets have been developed to address various challenges. VIPeR\cite{gray2007evaluating} is the first dataset focus on viewpoint invariant pedestrian recognition. PRID-2011\cite{hirzer2011person}, Market-1501\cite{zheng2015scalable} and MSMT17\cite{wei2018person} expanded on this with more person ID numbers and more camera views. Nevertheless, a single RGB modality struggles to cope with changes in lighting conditions. To address this, the infrared\cite{wu2017rgb} and thermal\cite{nguyen2017person} modalities have been introduced into ReID systems. Recently, several new cross-modal ReID datasets have been proposed. LLCM\cite{zhang2023diverse} focuses on the infrared pedestrian images across various illumination scenarios. HITSZ-VCM\cite{lin2022learning} has launched the first video-based infrared dataset. Additionally, CM-Group\cite{xiong2023similarity} has introduced the first cross-modality dataset specifically for group ReID. 
Lately, advancements of UAV has enabled ReID systems to overcome the limitations of stationary setups. MRP\cite{layne2015investigating} utilizes UAV as a moving platform to introduce the first dataset for UAV ReID system. DRHIT01\cite{grigorev2019deep} presents a small-scale dataset with real UAV view. For the subsequent works, PRAI-1581\cite{zhang2020person} emphasizes a variety of altitudes and  P-DESTRE\cite{kumar2020p} supports long-term ReID. Besides, UAV-Human\cite{li2021uav} propose a benchmark for human behavior understanding with UAV. More recently, AG-ReID \cite{nguyen2023aerial} and AG-ReID v2\cite{nguyen2024ag} have introduced a dataset across aerial RGB cameras and ground RGB cameras.
Nevertheless, extant research conducted on datasets has exclusively focused on data collected through CCTV or UAVs, and have only considered two modalities. There is a paucity of a dataset that can encompass a wider range of modalities and different collection platforms, which is necessary in complex urban environments and real-world applications.

\subsection{Multi-Modality Person Re-Identification}

Traditional person ReID primarily focuses on identifying pedestrians within single modality\cite{luo2019bag, he2020fastreid,hirzer2011person,li2014deepreid,wei2018person,zheng2015scalable,zheng2017unlabeled,Gao_2020_CVPR}, which struggles to overcome the limitations due to varying lighting conditions, typically performing inadequately in low lighting conditions. To address this problem, \cite{wu2017rgb,nguyen2017person} have introduced infrared and thermal modalities into the ReID system, which led to the emergence of numerous cross-modality ReID studies\cite{ye2021channel,ye2023channel, fang2023visible,ye2021deep,zhang2023diverse,wang2022optimal,zhang2022modality}. More specifically, CAJ\cite{ye2021channel}, CAJ$_+$ \cite{ye2023channel} and DEEN\cite{zhang2023diverse} have made efforts on augmentation strategies to narrow the gap between modalities. AGW\cite{ye2021deep}, MSCLNet\cite{zhang2022modality} and SAAI\cite{fang2023visible} have attempted to learn features that are invariant across modalities in a latent space. Additionally, to alleviate the laboring efforts, OTLA-ReID\cite{wang2022optimal} proposed an optimal transport based method to assign pseudo labels
from visible to infrared modality. 
However, existing methods focus solely on modality gaps between infrared/RGB or thermal/RGB, overlooking the challenges introduced by diverse capture devices—such as variations in viewpoints, resolutions, and indoor/outdoor environments. A unified framework addressing both cross-modality and cross-platform ReID is urgently needed.


\section{Dataset}
\subsection{Dataset Overview}

In urban environments, the inherent challenges—such as complex scenes, architectural occlusions, and high pedestrian mobility—demand the effective integration of diverse sensor modalities and capture platforms. Recognizing that these multi-modality and multi-platform configurations are essential yet underrepresented in existing benchmarks, we introduce MP-ReID, the first dataset tailored to address these practical challenges. 
Table.~\ref{tab:freq} provides a quantitative comparison between MP-ReID and several established ReID datasets, underscoring our dataset’s advantages in both scale and diversity of modalities and platforms. 
We collect images from different platforms and various modalities, making it particularly valuable for the development and evaluation of multi-modality multi-platform person ReID. 
The MP-ReID consists 1,930 identities and 136,156 images sampled from over 1.2M frames. The images are collected from ground RGB and infrared cameras and UAV RGB and thermal cameras. As illustrated in Fig.~\ref{fig:challenge}, our dataset comprises three distinct modalities and three different scenes, with notable disparities between images captured in different modalities and scenes, and it also illustrates the challenges present across various modalities and platforms.

\begin{table*}\centering
  \begin{tabular}{l|c|c|c|c|c|c|c|c|c|c} 
    \toprule
    \multirow{2}{*}{Dataset} & \multirow{2}{*}{IDs.} & \multirow{2}{*}{Cams.} & \multirow{2}{*}{BBox} & \multicolumn{3}{c|}{Scene}& \multicolumn{3}{c|}{Modality}&\multirow{2}{*}{Privacy}\\
    \cline{5-10}
    & & & & Indoor & Outdoor & Aerial & RGB & Infrared & Thermal\\
    \midrule
    PRID\cite{hirzer2011person}        & $200$  & 2 & $1,134$ &          &\checkmark&          &\checkmark&          &\\
    CUHK03\cite{li2014deepreid}        & $1,360$& 2 & $13,164$&          &\checkmark&          &\checkmark&          &\\ 
    Market-1501\cite{zheng2015scalable}  & $1,501$& 6 & $32,668$&          &\checkmark&          &\checkmark&          &\\
    MSMT17\cite{wei2018person}         & $4,101$& 15&$126,441$&          &\checkmark&          &\checkmark&          &\\
    \midrule
    \midrule
    PRAI-1581\cite{zhang2020person}    & $1,581$& 2 & $39,461$&          &\checkmark&\checkmark&\checkmark&          &\\
    UAV-Human\cite{li2021uav}          & $1,144$& 1 & $41,290$&          &\checkmark&\checkmark&\checkmark&          &\\
    P-DESTRE\cite{kumar2020p}             & $253$  & 1 & $>14.8$M&          &\checkmark&\checkmark&\checkmark&          &\\
    AG-ReID\cite{nguyen2023aerial}     & $388$  & 2 & $21,983$&          &\checkmark&\checkmark&\checkmark&          &  & \checkmark\\
    AG-ReID v2\cite{nguyen2024ag}     & $1,615$  & 3 & $100,502$&          &\checkmark&\checkmark&\checkmark&          &  & \checkmark\\
    G2APS\cite{zhang2023ground}     & $2,644$  & 2 & $260,559$&          &\checkmark&\checkmark&\checkmark&          &  & \\
    \midrule
    \midrule
    RegDB\cite{nguyen2017person}       & $412$  & 2 & $8,240$ &          &\checkmark&          &\checkmark&&\checkmark\\
    SYSU-MM01\cite{wu2017rgb}       & $491$  & 6 & $38,271$&\checkmark&\checkmark&          &\checkmark&\checkmark&\\
    LLCM\cite{zhang2023diverse}         & $1,064$& 9 & $46,767$&          &\checkmark&          &\checkmark&\checkmark&  &\checkmark\\
    HITSZ-VCM\cite{lin2022learning}      & $927$  & 12&$463,259$&\checkmark&\checkmark&          &\checkmark&\checkmark&\\
    \midrule
    \midrule
    MP-ReID (Ours)        & $1,930$& 14&$136,156$&\checkmark&\checkmark&\checkmark&\checkmark&\checkmark& 
    \checkmark & \checkmark\\
    \bottomrule
    \bottomrule

\end{tabular}
\caption{Comparing with Existing Person ReID Datasets. Existing datasets are organized into three distinct categories: (1) conventional RGB-based ReID datasets, (2) UAV ReID datasets, and (3) cross-modality ReID datasets.} 
  \label{tab:freq}
\end{table*}

\subsection{Comparison with Existing Person ReID Datasets.}Traditional datasets\cite{hirzer2011person,li2014deepreid,zheng2015scalable,zheng2017unlabeled,wei2018person} for pedestrian ReID tasks primarily encompass RGB images captured in outdoor environments. The exemplar among these, MSMT17\cite{wei2018person}, featuring 4,101 unique identities and a total of 126,441 bounding boxes, was acquired through 15 cameras.  Cross-modality datasets\cite{nguyen2017person,wu2017rgb,zhang2023diverse,lin2022learning} are devised for more complicated ReID tasks, often incorporating thermal or infrared modalities, and occasionally extending to indoor settings. Existing UAV datasets\cite{zhang2020person,li2021uav,kumar2020p,nguyen2023aerial,zhang2023ground,zhang2024cross} typically exhibit limited scope, with no more than three cameras deployed and restricted to RGB image acquisition for pedestrian ReID purposes.
However, our proposed MP-ReID dataset significantly extends beyond conventional limitations. Leveraging data collection from both ground-based cameras and UAV, our dataset contains video recordings capturing RGB, infrared, and thermal modalities across indoor, outdoor, and aerial domains.
Noteworthy is the scale of our MP-ReID dataset, featuring 1,930 distinct identities and 136,156 human bounding boxes acquired from 14 cameras. This scale is comparable to that of multi-view traditional datasets, thus establishing our dataset as a substantial resource for ReID research, featuring numerous cameras and human annotations across various platforms and modalities. To ensure robust privacy protection, we applied mosaic techniques to obscure pedestrians’ facial features in all video recordings and permanently deleted the original raw footage to prevent any potential data leakage.


\subsection{Data Collection}
To maximize dataset diversity and capture a wider range of scenes, we have employed a total of 12 Hikvision DS-2DE3Q140MX-T cameras, recording at a frame rate of 25 fps with a resolution of $1920 \times 1080$ for 1 hour. Notably, half of these cameras were configured to operate in full-color mode, while the remaining six were configured for infrared night vision mode. In addition to the ground-based cameras, a DJI Mavic 3T platform, which carried an RGB and a thermal camera placed in parallel with hardware synchronization, to conduct aerial data collection and capture pedestrian activities from an aerial perspective. The resolution of the RGB cameras is $3840 \times 2160$, whereas the resolution of the thermal camera is $640 \times 512$. Both cameras capture simultaneously at a frame rate of 30 fps. Furthermore, the cruising altitudes were set at 5m, 7m, and 10m, capturing videos from varying perspectives and angles ranging from 30 to 80 degrees relative to the ground. The total duration of the collected videos exceeded 13 hours.
Our research team has received ethics approval from the Institutional Review Board for this project and the building administration has authorized the recording. All participants have been notified that data collection is solely for research objectives. To enhance privacy protection, we have applied facial blurring to all video recordings and permanently deleted the raw footage.

\begin{figure*}[htbp]
  \centering
  \includegraphics[width=0.75\linewidth]{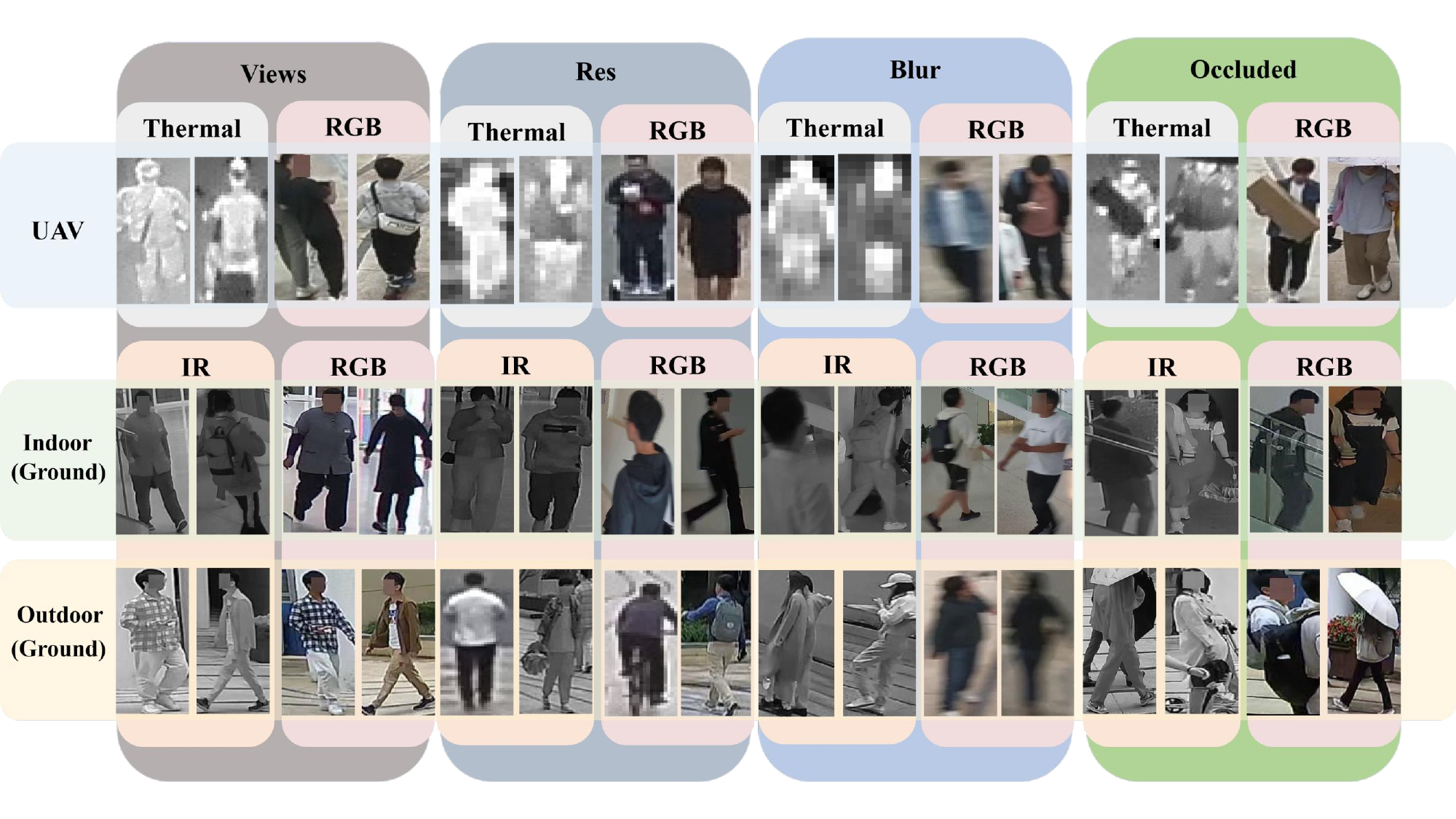}
  \caption{
 Our MP-ReID dataset comprises three distinct modalities and three different scenes, with notable disparities between images captured in different modalities and scenes. We showcase a range of variations to highlight the challenges present in person re-identification. From left to right, selected samples from different scenes and modalities illustrate the disparities between various viewpoints, instances of low resolution, cases of motion blur, and scenarios involving occlusion, respectively. These examples serve to demonstrate the complex nature of the gaps and obstacles within our dataset, emphasizing the diversity and real-world applicability of the MP-ReID benchmark.}
  \label{fig:challenge}
\end{figure*}

\subsection{Privacy Preserving}

Ethical approval for this study was obtained from the Institutional Review Board. We secured all necessary shooting permits for the designated areas and ensured compliance by posting public notices within the shooting zones. All cameras were placed in prominent locations for transparency. Participants entering the shooting zones were informed of the protocol and provided explicit consent. To safeguard data privacy, we meticulously anonymized the released dataset by applying mosaic techniques to obscure pedestrians’ facial features. Furthermore, we permanently deleted the raw footage, retaining only the processed dataset to eliminate any risk of privacy exposure. To ensure responsible use, the dataset will be restricted to non-commercial academic purposes, enforced through signed agreements that explicitly prohibit commercial exploitation


\subsection{Data Annotation}
To ensure the acquisition of high-quality pedestrian bounding boxes, we utilized the state-of-the-art YOLOX\cite{ge2021yolox} method to track the aforementioned videos. In total, we obtained over 1,500,000 pedestrian bounding boxes. Furthermore, we employed manual filtering on the obtained bounding boxes, resulting in over 790,000 usable bounding boxes from ground cameras and over 250,000 from UAV's RGB camera. The process of aligning identities obtained from disparate camera sources necessitated a substantial investment of time and effort. In particular, we found that due to the low resolution and significant modal differences of the UAV's thermal camera, the tracking performance of YOLOX\cite{ge2021yolox} in thermal videos was unsatisfactory. To overcome this limitation, for thermal videos, we manually compared each frame with the annotated UAV's RGB videos and manually delineated bounding boxes. Ultimately, this rigorous procedure yielded 1,930 unique IDs, ensuring the accuracy and reliability of our dataset. 

\subsection{Data Statistics}
In our dataset, each ground RGB camera captures an average of 497 IDs and 7,545 bounding boxes, each ground infrared camera captures an average of 369 IDs and 6,050 bounding boxes, the UAV's RGB camera captures 341 IDs and 26,046 bounding boxes, and the UAV's thermal camera captures 474 IDs and 28,543 bounding boxes. There are 346 persons captured by only one camera and 381 persons with less than 10 bounding boxes, most person are captured by 2-8 cameras and have 10-60 bounding boxes.
Since outdoor scenes pose significant challenges, such as varying lighting conditions, occlusions, and high pedestrian densities, making them essential for robust ReID performance, it is reasonable to have a larger proportion of outdoor scene data than indoor scenes.
The bounding box ratios of persons in outdoor, indoor and aerial view are $51.2\%$, $8.8\%$ and $40.1\%$, the bounding box ratios of persons in RGB, infrared and thermal modality are $52.3\%$, $26.7\%$ and $21.0\%$.
 For the test IDs appearing in one modality or platform, we randomly select one image from each camera as the query, while all data from the other modality or platform are used as the gallery during testing. 
More statistical details and the data splitting details can be found in the Supplementary Material.


\subsection{Limitations}
\label{sec:limit}
While MP-ReID includes data from multiple modalities and platforms, there are still additional modalities and platforms that could be incorporated, such as the wearable device platforms\cite{nguyen2024ag} and the event camera data\cite{ahmad2023person}. We encourage future researchers to supplement our work to some extent.

\section{Uni-Prompt ReID}

To enhance cross-modality and cross-platform generalization in ReID, we introduce Uni-Prompt ReID, a unified prompt-based learning framework that leverages vision-language alignment in VLMs like CLIP through multi-part text prompt fine-tuning. Building on DAPrompt \cite{DAPrompt}, we incorporate modality and platform information into the text prompt. Additionally, inspired by CoCoOp\cite{CoCoOp}, we enhance learnable prompts by designing a weight-tied meta-network to extract text prompt biases from pedestrian image visual features.

\begin{figure*}[h]
    
    \centering
    \includegraphics[width=0.75\linewidth]{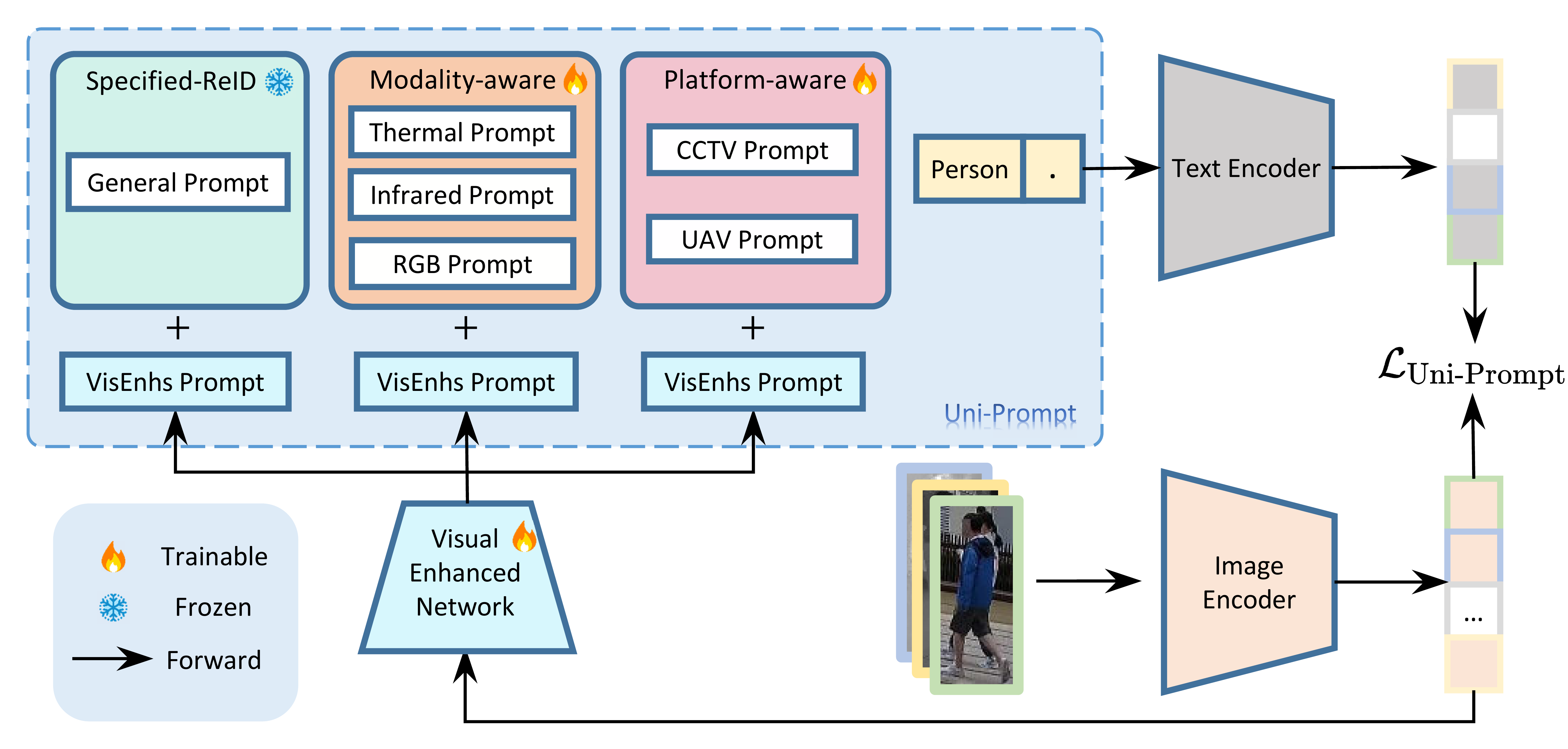}
    \caption{Uni-Prompt ReID divides the learnable context into three parts: Specified-ReID Prompt, Modality-Aware Prompt, and Platform-Aware Prompt. During the training stage, the specified-ReID prompt will be learned first in a warm-up stage and then be frozen with the update of modality-aware prompt and platform-aware prompt. An Inherent Information Embedding Net is also updated with the prompts. }
    \label{fig:mainfig}
    
\end{figure*}

Our framework employs a learnable text prompt structure, as illustrated in Fig.~\ref{fig:mainfig}, which integrates multi-modality and multi-platform information into prompt learning. The text prompt comprises three distinct components: (1) the Specific Re-ID Prompt, which encodes individual-specific information; (2) the Modality-aware Prompt, which captures modality-specific details; and (3) the Platform-aware Prompt, which incorporates platform-specific context. This design enhances the model’s representational capacity by embedding rich, task-relevant information into the prompt structure.

Moreover, image features contain inherent information that can assist in distinguishing prompts during learning. For example, an infrared image feature may help guide the modality-aware prompt closer to an infrared-specific representation.  To use the visual information for enhancing the learnable text prompt, we design a lightweight neural network, named visual-enhanced network, which embeds the image feature into context vectors as a conditioning factor. Let visual-enhanced network $g_{\theta}(\cdot)$ be parameterized by $\theta$. Given the image feature $a$, the visual-enhanced context vector is defined as $\sigma :=(\sigma_X, \sigma_P,\sigma_M) = g_{\theta}(a)$. The visual-enhanced text prompt is then defined as $S_m(a) = [S]_m + \sigma_S$ where $[S]_m\in\{X_m, P_m,M_m\}$ represents the $m$-th text token before enhancement. Here, $X, P, M$ denote specific ReID prompts, modality-aware prompts, and platform-aware prompts, respectively. The visual-enhanced context vector $\sigma$ is added to the corresponding text prompt to enhance the modality and platform context. During training, the prompt vectors are updated along with the VE-Net parameters \( \theta \). Overall, the $i$-th person prompt is then formulated as:
\begin{equation}
    \begin{aligned}
    t_i(a) = &X_1(a)X_2(a)\dots X_M(a)\\ &P_1(a)P_2(a)\dots P_R(a)\\ &M_1(a)M_2(a)\dots M_B(a), \text{person}_i.
\end{aligned}
\end{equation}

During the text prompt learning phase, we use a two-stage training process. In the first stage, we fix the modality-aware prompt and platform-aware prompt and use image to text loss and text to image loss proposed in  $\mathcal{L}_{i2t}$ and $\mathcal{L}_{t2i}$ from CLIP-ReID \cite{CLIP-ReID} to learn the specific Re-ID prompt. 

In the second stage, we fix the learned specified Re-ID prompt and use $\mathcal{L}_{m}$ and $\mathcal{L}_{p}$ to learn the modality-specific and platform-specific prompt. The modality-specific and platform-specific contrastive losses are separately calculated as:
\begin{equation}
    \mathcal{L}_{mi2t}(i)=-\log\frac{\exp(s(V_i,{T_{m}}_i))}{\sum_{a=1}^B\exp(s(V_a,{T_{m}}_a))}
\end{equation}
\begin{equation}
\mathcal{L}_{mt2i}(c_i)=\frac{-1}{|P(c_i)|}\sum_{p\in P(c_i)}\log\frac{\exp(s(V_p,{T_m}_{c_i}))}{\sum_{a=1}^B\exp(s(V_a,{T_m}_{c_i}))}
\end{equation}
\begin{equation}
    \mathcal{L}_{pi2t}(i)=-\log\frac{\exp(s(V_i,{T_{p}}_i))}{\sum_{a=1}^B\exp(s(V_a,{T_{p}}_a))}
\end{equation}
\begin{equation}
    \mathcal{L}_{pt2i}(c_i)=\frac{-1}{|P(c_i)|}\sum_{u\in P(c_i)}\log\frac{\exp(s({V}_u,{T_p}_{c_i}))}{\sum_{a=1}^B\exp(s({V}_a,{T_p}_{c_i}))} 
\end{equation}
where the ${T_m}$ and ${T_{p}}$ are embedded text features with fixed vanilla tokens and specific modality or platform tokens. The number of positive samples is defined as $P(c_i)=\{p\in1...B:y_{p}=y_{i}\}$. The overall loss for Uni-Prompt is expressed as:
\begin{equation}
    \mathcal{L}_{\text{Uni-Prompt}}= \mathcal{L}_{mi2t}+\mathcal{L}_{mt2i}+\mathcal{L}_{pi2t}+\mathcal{L}_{pt2i}
\end{equation}

\section{Experiments}
\label{sec:blind}

\subsection{Evaluation Metrics}

Following the work of \cite{eom2019learning,zheng2015scalable,wei2018person}, we employed both Cumulated Matching Characteristic (CMC) at Rank-1 and Rank-5. To account for the retrieval of multiple instances and difficult samples, we use mean average precision (mAP) as the accuracy metrics.

\subsection{Implementation Details}
We established the benchmark by employing our Uni-Prompt ReID method alongside several representative baseline methods in the field of Visible-Infrared ReID \cite{ye2021channel,ye2023channel,ye2021deep,zhang2023diverse,wang2022optimal,fang2023visible,CSDN}. Details of these baselines are provided in the next subsection. To ensure a fair comparison, we conducted experiments on a cluster equipped with eight Nvidia RTX 4090 GPUs each with 24GB of memory. During training, we employed random erasing with a probability of 0.5, random horizontal flipping, and random cropping for data augmentation. All other configurations were set to their default values. For testing, we randomly selected a query image for each identity per camera, and all the images of identities captured by the corresponding camera were used as the gallery. Testing for each model and setting was repeated across 10 trials, and the average metrics were reported. Further details on the training and testing data splitting can be found in Supplementary Materials.




\begin{table*}[h]
\centering
\resizebox{\linewidth}{!}{
\begin{tabular}{ccccccccccccccc}
\toprule
\hline
\multicolumn{3}{c}{\multirow{2}{*}{}}   & \multicolumn{3}{c|}{Cross-platform Only} & \multicolumn{3}{c|}{Cross-modality Only}  & \multicolumn{3}{c|}{Cross-modality \& platform} & \multicolumn{3}{c}{Average}  \\
\multicolumn{3}{c}{}                                          & \multicolumn{1}{c}{Rank-1} & \multicolumn{1}{c}{Rank-5}  & \multicolumn{1}{c|}{mAP} & \multicolumn{1}{c}{Rank-1} & \multicolumn{1}{c}{Rank-5}  & \multicolumn{1}{c|}{mAP} & \multicolumn{1}{c}{Rank-1} & \multicolumn{1}{c}{Rank-5}  & \multicolumn{1}{c|}{mAP} & \multicolumn{1}{c}{Rank-1} & \multicolumn{1}{c}{Rank-5}  & \multicolumn{1}{c}{mAP}
\\ \hline

\multicolumn{2}{l}{CAJ~\cite{ye2021channel}}               &  & \multicolumn{1}{c}{40.36}  & \multicolumn{1}{c}{63.81} & \multicolumn{1}{c|}{29.77}     & \multicolumn{1}{c}{45.34}  & \multicolumn{1}{c}{64.63} & \multicolumn{1}{c|}{26.84} & \multicolumn{1}{c}{10.62}  & \multicolumn{1}{c}{24.27} & \multicolumn{1}{c|}{7.91} & \multicolumn{1}{c}{32.11}  & \multicolumn{1}{c}{50.90} & \multicolumn{1}{c}{21.51} \\ 

\multicolumn{2}{l}{\multirow{1}{*}{CAJ$_+$~\cite{ye2023channel}}}               &  & \multicolumn{1}{c}{47.60}  & \multicolumn{1}{c}{70.09} & \multicolumn{1}{c|}{37.53}     & \multicolumn{1}{c}{58.16}  & \multicolumn{1}{c}{76.17} & \multicolumn{1}{c|}{38.52} & \multicolumn{1}{c}{21.51}  & \multicolumn{1}{c}{41.61} & \multicolumn{1}{c|}{15.79}  & \multicolumn{1}{c}{42.42}  & \multicolumn{1}{c}{62.62} & \multicolumn{1}{c}{30.61} \\

\multicolumn{2}{l}{\multirow{1}{*}{AGW}~\cite{ye2021deep}}               &  & \multicolumn{1}{c}{53.68}  & \multicolumn{1}{c}{75.88} & \multicolumn{1}{c|}{42.88}     & \multicolumn{1}{c}{51.88}  & \multicolumn{1}{c}{69.32} & \multicolumn{1}{c|}{33.90} & \multicolumn{1}{c}{19.21}  & \multicolumn{1}{c}{37.94} & \multicolumn{1}{c|}{14.91}& \multicolumn{1}{c}{41.59}  & \multicolumn{1}{c}{61.05} & \multicolumn{1}{c}{30.56} \\

\multicolumn{2}{l}{\multirow{1}{*}{DEEN}~\cite{zhang2023diverse}}               &  & \multicolumn{1}{c}{60.05}  & \multicolumn{1}{c}{79.82} & \multicolumn{1}{c|}{49.02}     & \multicolumn{1}{c}{69.59}  & \multicolumn{1}{c}{\textbf{85.55}} & \multicolumn{1}{c|}{48.88} & \multicolumn{1}{c}{27.59}  & \multicolumn{1}{c}{51.76} & \multicolumn{1}{c|}{20.07}  & \multicolumn{1}{c}{52.41}  & \multicolumn{1}{c}{72.38} & \multicolumn{1}{c}{39.33}  \\  

\multicolumn{2}{l}{\multirow{1}{*}{OTLA-ReID}~\cite{wang2022optimal}}               &  & \multicolumn{1}{c}{73.24}  & \multicolumn{1}{c}{85.60} & \multicolumn{1}{c|}{61.85}     & \multicolumn{1}{c}{68.12}  & \multicolumn{1}{c}{84.72} & \multicolumn{1}{c|}{46.57} & \multicolumn{1}{c}{29.31}  & \multicolumn{1}{c}{53.27} & \multicolumn{1}{c|}{20.66}  & \multicolumn{1}{c}{56.89}  & \multicolumn{1}{c}{74.53}  & \multicolumn{1}{c}{43.03} \\

\multicolumn{2}{l}{\multirow{1}{*}{SAAI}~\cite{fang2023visible}}               &  & \multicolumn{1}{c}{68.11}  & \multicolumn{1}{c}{82.71} & \multicolumn{1}{c|}{53.50}     & \multicolumn{1}{c}{60.42}  & \multicolumn{1}{c}{77.98} & \multicolumn{1}{c|}{41.71} & \multicolumn{1}{c}{28.40}  & \multicolumn{1}{c}{51.59} & \multicolumn{1}{c|}{20.31} & \multicolumn{1}{c}{52.31}   & \multicolumn{1}{c}{70.76} & \multicolumn{1}{c}{38.51}  \\  

\multicolumn{2}{l}{\multirow{1}{*}{CSDN}~\cite{CSDN}}               &  & \multicolumn{1}{c}{31.65}  & \multicolumn{1}{c}{54.16} & \multicolumn{1}{c|}{23.22}     & \multicolumn{1}{c}{41.34}  & \multicolumn{1}{c}{53.90} & \multicolumn{1}{c|}{32.73} & \multicolumn{1}{c}{12.17}  & \multicolumn{1}{c}{20.49} & \multicolumn{1}{c|}{11.40} & \multicolumn{1}{c}{28.39}   & \multicolumn{1}{c}{42.85} & \multicolumn{1}{c}{22.45}  \\ 

\multicolumn{2}{l}{\multirow{1}{*}{\textbf{Uni-Prompt(Ours)}}}               &  & \multicolumn{1}{c}{\textbf{78.77}}  & \multicolumn{1}{c}{\textbf{88.38}} & \multicolumn{1}{c|}{\textbf{74.29}}     & \multicolumn{1}{c}{\textbf{72.26}}  & \multicolumn{1}{c}{85.00} & \multicolumn{1}{c|}{\textbf{60.81}} & \multicolumn{1}{c}{\textbf{43.16}}  & \multicolumn{1}{c}{\textbf{60.60}} & \multicolumn{1}{c|}{\textbf{40.26}}  & \multicolumn{1}{c}{\textbf{64.73}}  & \multicolumn{1}{c}{\textbf{77.99}}  & \multicolumn{1}{c}{\textbf{58.45}} \\  


\hline
\bottomrule
\end{tabular}
}
\caption{We classified the experimental settings into three categories: cross-modality only, cross-platform only, and both cross-modality and cross-platform. We then compare the average performance of these three categories.}
\label{tab:categoried results}
\end{table*}

\subsection{Experimental Results}
To establish a comprehensive benchmark, we compared our Uni-Prompt ReID method with the following baseline approaches.
\textbf{CAJ}~\cite{ye2021channel} employs a channel-augmented joint learning strategy for color robustness and an enhanced channel-mixed learning approach to address modality variations.
 \textbf{CAJ$_+$}~\cite{ye2023channel}  utilizes  random channel exchange and a weak-and-strong augmentation joint learning strategy, significantly enhancing its effectiveness.
 \textbf{AGW}~\cite{ye2021deep} incorporates a non-local attention block \cite{yu2017cross} and generalized-mean pooling \cite{radenovic2018fine} with weighted triplet loss for cross-modality visible-infrared tasks. \textbf{DEEN}~\cite{zhang2023diverse} proposes a diverse embedding expansion module with a center-guided pair mining loss for data augmentation in the embedding space to narrow gaps between visible and infrared images. \textbf{OTLA-ReID}~\cite{wang2022optimal} proposes an Optimal-Transport label assignment module to align infrared and RGB images and employs prediction alignment learning to mitigate the negative effects of inaccurate pseudo-labels.
 \textbf{SAAI}~\cite{fang2023visible} aggregates latent features via pixel-prototype correlation and optimizes inference with a pedestrian affinity module.
 \textbf{CSDN}~\cite{CSDN} proposes a CLIP-driven semantic discovery network to integrate text features from both visible and infrared modalities, embedding comprehensive semantics into visual representation. 

 To validate the various characteristics of our proposed MP-ReID, we designed a total 12 different experimental settings, considering all four image sources: ground RGB, ground infrared, UAV RGB and UAV thermal. It is important to note that, in order to address the complexity and variability of city-scale scenarios, it is essential to fully leverage the interaction between multiple platforms and the complementarity of different modalities. To this end, we propose four novel cross-modality and cross-platform ReID settings. The results from each experiment were recorded and subsequently categorized into cross-modality, cross-platform, and cross-modality $\&$ platform. The mean performance of each method in each experimental category is shown in Table.~\ref{tab:categoried results}. The findings in the table reveal that most of the cross-modality methods perform well in cross-modality tasks but underperform in the cross modality $\&$ platform settings due to the lack of consideration for platform diversity. Our Uni-Prompt ReID method outperforms other state-of-the-art methods, with a 7.87$\%$ boost in average Rank-1 accuracy and a 15.42$\%$ higher mAP by on our MP-ReID dataset, demonstrating that Uni-Prompt ReID method exhibits strong robustness and high efficiency.  Further details of all 12 experimental results are provided in the Supplementary Materials. 

\begin{table*}[h]
\centering
\resizebox{\linewidth}{!}{
\begin{tabular}{ccccccccccccccc}
\toprule
\hline
\multicolumn{3}{c}{\multirow{2}{*}{}}   & \multicolumn{3}{c|}{Cross-platform Only} & \multicolumn{3}{c|}{Cross-modality Only}  & \multicolumn{3}{c|}{Cross-modality \& platform} & \multicolumn{3}{c}{Average}  \\
\multicolumn{3}{c}{}                                          & \multicolumn{1}{c}{Rank-1} & \multicolumn{1}{c}{Rank-5}  & \multicolumn{1}{c|}{mAP} & \multicolumn{1}{c}{Rank-1} & \multicolumn{1}{c}{Rank-5}  & \multicolumn{1}{c|}{mAP} & \multicolumn{1}{c}{Rank-1} & \multicolumn{1}{c}{Rank-5}  & \multicolumn{1}{c|}{mAP} & \multicolumn{1}{c}{Rank-1} & \multicolumn{1}{c}{Rank-5}  & \multicolumn{1}{c}{mAP}
\\ \hline

\multicolumn{2}{l}{Base}               &  & \multicolumn{1}{c}{77.01}  & \multicolumn{1}{c}{81.63} & \multicolumn{1}{c|}{71.19}     & \multicolumn{1}{c}{61.11}  & \multicolumn{1}{c}{70.71} & \multicolumn{1}{c|}{49.68} & \multicolumn{1}{c}{28.40}  & \multicolumn{1}{c}{38.53} & \multicolumn{1}{c|}{23.08} & \multicolumn{1}{c}{55.51}  & \multicolumn{1}{c}{63.62} & \multicolumn{1}{c}{47.98} \\ 

\multicolumn{2}{l}{\multirow{1}{*}{Base+MP}}               &  & \multicolumn{1}{c}{77.18}  & \multicolumn{1}{c}{84.26} & \multicolumn{1}{c|}{71.34}     & \multicolumn{1}{c}{67.34}  & \multicolumn{1}{c}{81.24} & \multicolumn{1}{c|}{56.77} & \multicolumn{1}{c}{31.57}  & \multicolumn{1}{c}{46.18} & \multicolumn{1}{c|}{26.91}  & \multicolumn{1}{c}{58.70}  & \multicolumn{1}{c}{70.56} & \multicolumn{1}{c}{51.67} \\

\multicolumn{2}{l}{\multirow{1}{*}{Base+MP+PP}}               &  & \multicolumn{1}{c}{78.62}  & \multicolumn{1}{c}{88.01} & \multicolumn{1}{c|}{73.68}     & \multicolumn{1}{c}{70.31}  & \multicolumn{1}{c}{81.88} & \multicolumn{1}{c|}{60.19} & \multicolumn{1}{c}{40.66}  & \multicolumn{1}{c}{56.55} & \multicolumn{1}{c|}{38.57}& \multicolumn{1}{c}{63.20}  & \multicolumn{1}{c}{75.48} & \multicolumn{1}{c}{57.48} \\  

\multicolumn{2}{l}{\multirow{1}{*}{Base+MP+PP+VE(Full)}}               &  & \multicolumn{1}{c}{\textbf{78.77}}  & \multicolumn{1}{c}{\textbf{88.38}} & \multicolumn{1}{c|}{\textbf{74.29}}     & \multicolumn{1}{c}{\textbf{72.26}}  & \multicolumn{1}{c}{\textbf{85.00}} & \multicolumn{1}{c|}{\textbf{60.81}} & \multicolumn{1}{c}{\textbf{43.16}}  & \multicolumn{1}{c}{\textbf{60.60}} & \multicolumn{1}{c|}{\textbf{40.26}}  & \multicolumn{1}{c}{\textbf{64.73}}  & \multicolumn{1}{c}{\textbf{77.99}}  & \multicolumn{1}{c}{\textbf{58.45}} \\  

\hline
\bottomrule
\end{tabular}
}
\caption{Ablation study of specifically designed prompts. Here, the Base is the specific ReID prompt learned by the CLIP-ReID loss, and MP, PP, VE represent the modality-aware prompt, platform-aware prompt and visual-enhanced prompt, respectively.}
\label{ablation1}
\end{table*}

\subsection{Ablation Study}
We conducted ablative analysis on Uni-Prompt ReID using the MP-ReID dataset, categorizing results into cross-modality, cross-platform, and cross-modality \&platform settings. Table.~\ref{ablation1} confirms its effectiveness in cross-modality \& platform scenarios, with detailed results in Supplementary Materials.


 \textbf{Modality-aware Prompt.} We conducted a comparative analysis of the Modality-aware prompt and a naive prompt, referred to as the base in Table.~\ref{ablation1}, which incorporates learnable specific-ReID tokens. Specifically, an average increase of 6.23$\%$ in Rank-1 accuracy and a notable 7.09$\%$ increase in mAP were observed for cross-modality settings. Additionally, the performance of both cross-platform settings and cross-modality $\&$ platform settings was marginally enhanced. These results demonstrate that the Modality-aware prompt significantly improves accuracy in cross-modality tasks without negatively impacting performance in other settings.


\textbf{Platform-aware Prompt.}We enhanced the Modality-aware prompt to a Modality $\&$ Platform-aware prompt by adding a learnable platform-aware prompt, improving Rank-1 accuracy and mAP in both cross-platform and cross-modality $\&$ platform tasks. Notably, this update surpassed state-of-the-art methods, especially in complex cross-modality $\&$ platform scenarios, highlighting the prompt’s effectiveness.

\textbf{Visual-enhanced Prompt.} As demonstrated in Table.~\ref{ablation1}, the inclusion of the Visual-enhanced Prompt into the Modality $\&$ Platform-aware prompt mentioned above leads to further improvement in each experiment and brings a 1.85\% Rank-1 accuracy boost in Cross-modality only experiments and a 2.50\% increase in those of the cross-modality $\&$ platform experiments. What's more, Uni-Prompt ReID achieves state-of-the-art performance within the visual-enhanced prompt, highlighting the advantage of this method in the complex scenarios where inconsistencies in visual features across modalities and differences in camera features in cross-platform scenes coexist.

\subsection{Result Analysis} 


As shown in Table.~\ref{tab:categoried results}, we categorized the total experimental settings into three classes: cross-modality only, cross-platform only, and both cross-modality and cross-platform. A general observation from our results indicates that existing baseline methods struggle to achieve satisfactory performance across these different settings within our dataset. The only scenarios where their performance can be considered acceptable are those that involve either a single cross-modality or a single cross-platform setting, provided that ground camera RGB data is available. This relatively better performance in such cases can be attributed to the smaller disparity introduced when only one type of variation is present. Additionally, the RGB data from ground cameras offers rich visual information, which serves as a crucial contextual aid in mitigating the challenges posed by either cross-modality or cross-platform discrepancies.

However, when it comes to the scenarios that include both cross-modality and cross-platform simultaneously, existing baseline methods exhibit significant shortcomings. This sharp decline is due to the fact that existing ReID datasets do not encompass the highly challenging scenarios as shown in Fig.~\ref{fig:challenge}, involving simultaneous cross-modality and cross-platform variations.

This performance gap highlights the necessity of our proposed MP-ReID benchmark, designed to overcome limitations in current datasets and evaluation settings. Unlike conventional methods, our Uni-Prompt ReID approach unifies multiple prompt mechanisms to handle both cross-modal and cross-platform variations. Specifically, it integrates modality-aware, platform-aware, and visual-enhanced prompts within a well-structured design, effectively capturing and leveraging information across modalities and platforms. This leads to superior performance across multiple tasks, setting new SOTA results and offering a promising solution to real-world multi-modal, multi-platform ReID challenges.

\section{Conclusion}

In this work, we introduce the MP-ReID dataset, a comprehensive multi-modality, multi-platform ReID benchmark designed to address the limitations of existing datasets in capturing modality and device variations.
Compared to previous datasets, MP-ReID offers a broader range of modalities and platforms, making it one of the most extensive benchmarks available. 
Its diverse variations—including differences in viewpoints, lighting, resolutions, and environments—ensure a more challenging and realistic evaluation for real-world ReID tasks. To tackle these challenges, we propose Uni-Prompt ReID, a unified framework specifically designed for cross-modality and cross-platform ReID. Our approach not only achieves state-of-the-art performance across all experimental setups but also significantly improves the handling of intricacies and domain gaps in multi-modal, multi-platform ReID.

{
    \small
    \bibliographystyle{ieeenat_fullname}
    \bibliography{main}
}







\appendix

%


\maketitlesupplementary




\begin{figure*}[h]
    \centering
    \includegraphics[width=\linewidth]{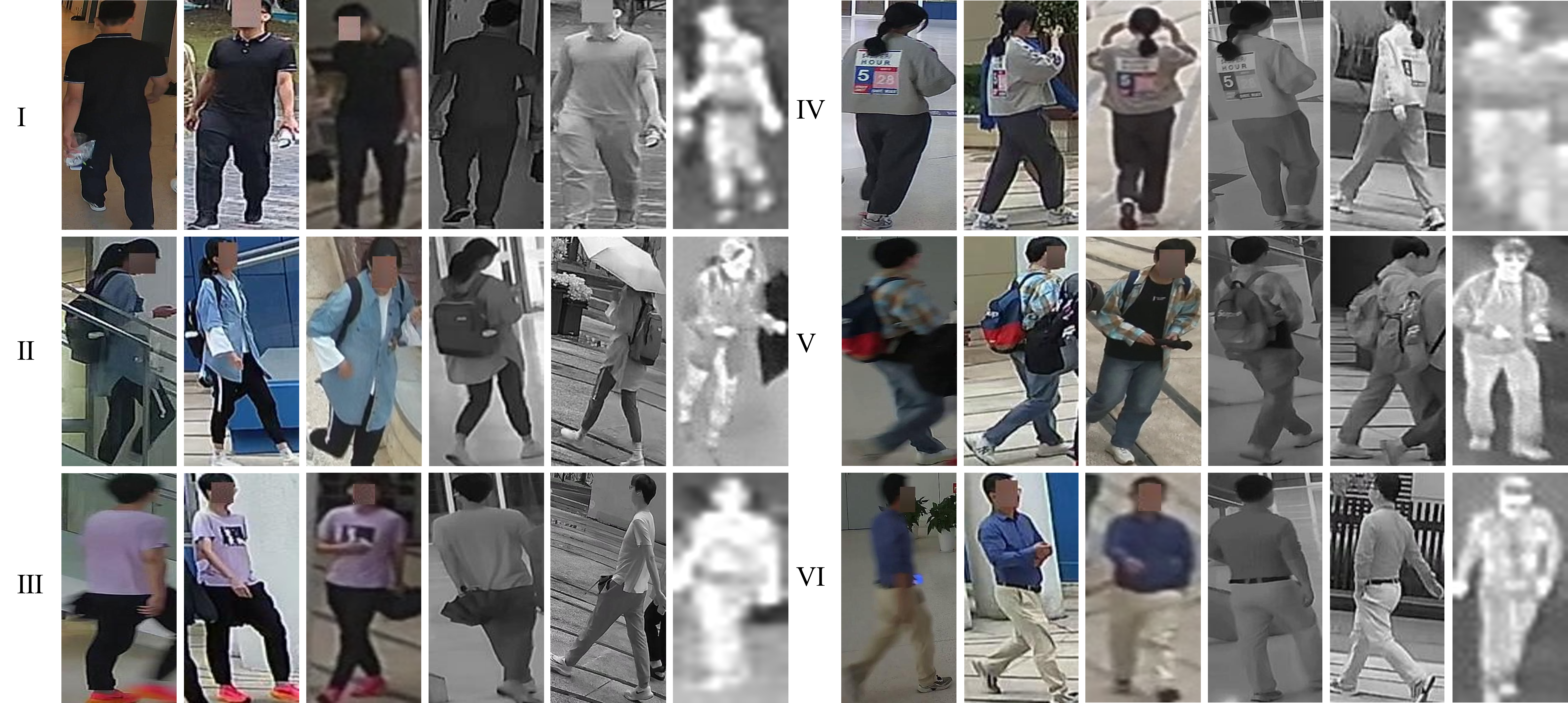}
    \caption{We provide 6 examples showing images of the same IDs in different scenes and various modalities within our MP-ReID. From left to right, indoor RGB, outdoor RGB, UAV RGB, indoor infrared, outdoor infrared and UAV thermal are shown for each ID, respectively.}
    \label{fig:more-examples}
\end{figure*}

\section{Distractors}

In the real-world scene, some individuals may only appear in one camera view, and we designate these as distractors. Table \ref{tab:distractor results} illustrates the performance of each baseline method after incorporating distractors, where the number of distractors is set to 10\% of the gallery. Here, we only aim to show the impact of distractors on the results, which is not the primary focus of our work. Therefore, we have selected experimental results from six representative settings.
Naturally, the presence of distractors inevitably impacts the performance of models, leading to varying degrees of degradation across different settings for each baseline method.

\begin{table*}[h]\centering
  \caption{\textbf{Dataset spliting of MP-ReID.} $\text{U}_T$, $\text{U}_R$, $\text{G}_I$ and $\text{G}_R$ stand for UAV thermal, UAV RGB, ground infrared and ground RGB, respectively. In the \textbf{Test} part, the number on the left of '/' corresponds to '$\rightarrow$' direction in the \textbf{Setting} part and the right corresponds to '$\leftarrow$'. And $\text{BBOX}_\text{q}$, $\text{BBox}_\text{g}$ exhibits the division of query and gallery bounding boxes.
  }
  \label{tab:split}
  \centering
  \begin{tabular}{c|c|c|c|c|c|c|c}
    \toprule
    \multicolumn{2}{c|}{Setting} & $\text{U}_{T} \leftrightarrow\text{G}_{R}$ &
              $\text{U}_{T} \leftrightarrow \text{U}_{R}$ &
              $\text{U}_{T} \leftrightarrow \text{G}_{I}$ &
              $\text{G}_{I} \leftrightarrow \text{U}_{R}$ & 
              $\text{G}_{I} \leftrightarrow \text{G}_{R}$ &
              $\text{G}_{R} \leftrightarrow \text{U}_{R}$ \\
    \midrule
    \multirow{2}{*}{Train}& ID     & 312 & 226 & 274 & 197 & 501 & 227 \\
    & BBox     & 29,670 & 35,563 & 24,308 & 20,045 & 26,265 & 24,978 \\
    \midrule
    \multirow{4}{*}{Test} &   ID  & 155 & 112 & 136 &  98 &1,001 & 113 \\
                          & $\text{BBox}_\text{q}$ & 155/329  &112/112  &136/235  &177/98    & 1,481/1,689 &239/113   \\ 
                          &$\text{BBox}_\text{g}$&4,940/8,373   &7,266/7,056   &3,777/7,957   &8,613/3,109  & 26,815/23,852 &8,600/3,273   \\
                       &$\text{BBox}_\text{Distractor}$&30427/402   &14/4690   &24826/3973   &  2712/27865  & 4393/283 & 17/34447  \\
  \bottomrule
  \end{tabular}
\end{table*}

\section{Details of MP-ReID}

Here, we show some details of the MP-ReID datasets, including the training and testing data splitting and some statistics analysis.
\subsection{Data Splitting}
We designed six sets of experiments for ground RGB, ground infrared, UAV RGB, UAV thermal data. For each set, we shuffle pedestrians simultaneously captured by two modal sensors or platforms and randomly select about 2/3 of the IDs as the training set. The remaining part is used as the test set, and the pedestrians captured by only one modal sensor or platform are used as distractors. 
One exception is that, due to the large volume of ground RGB and ground infrared data and the expensive data annotation in real scenarios, we only randomly select 1/3 as the training set in $\text{G}_{R} \leftrightarrow \text{G}_{I}$. For the test IDs appearing in one modality or platform, we randomly select one image from each camera as the query, while all data from the other modality or platform are used as the gallery during testing.

\begin{figure*}[h]
  \centering
  \begin{subfigure}{0.45\linewidth}
    \includegraphics[width=\linewidth]{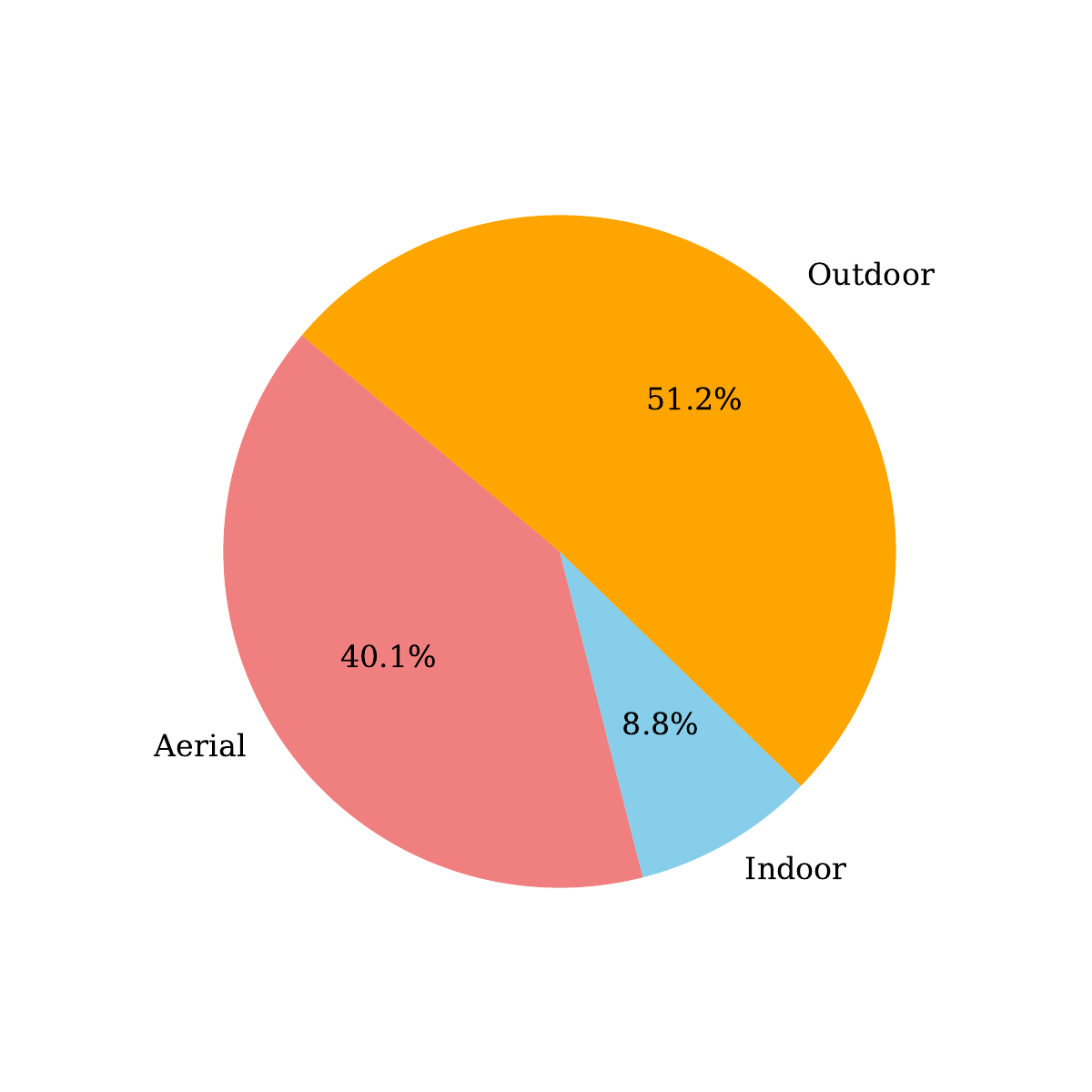}
    \caption{The proportion of bounding boxes in different scenes.}
    \label{fig:pie-a}
  \end{subfigure} 
  \hfill 
  \begin{subfigure}{0.45\linewidth}
    \includegraphics[width=\linewidth]{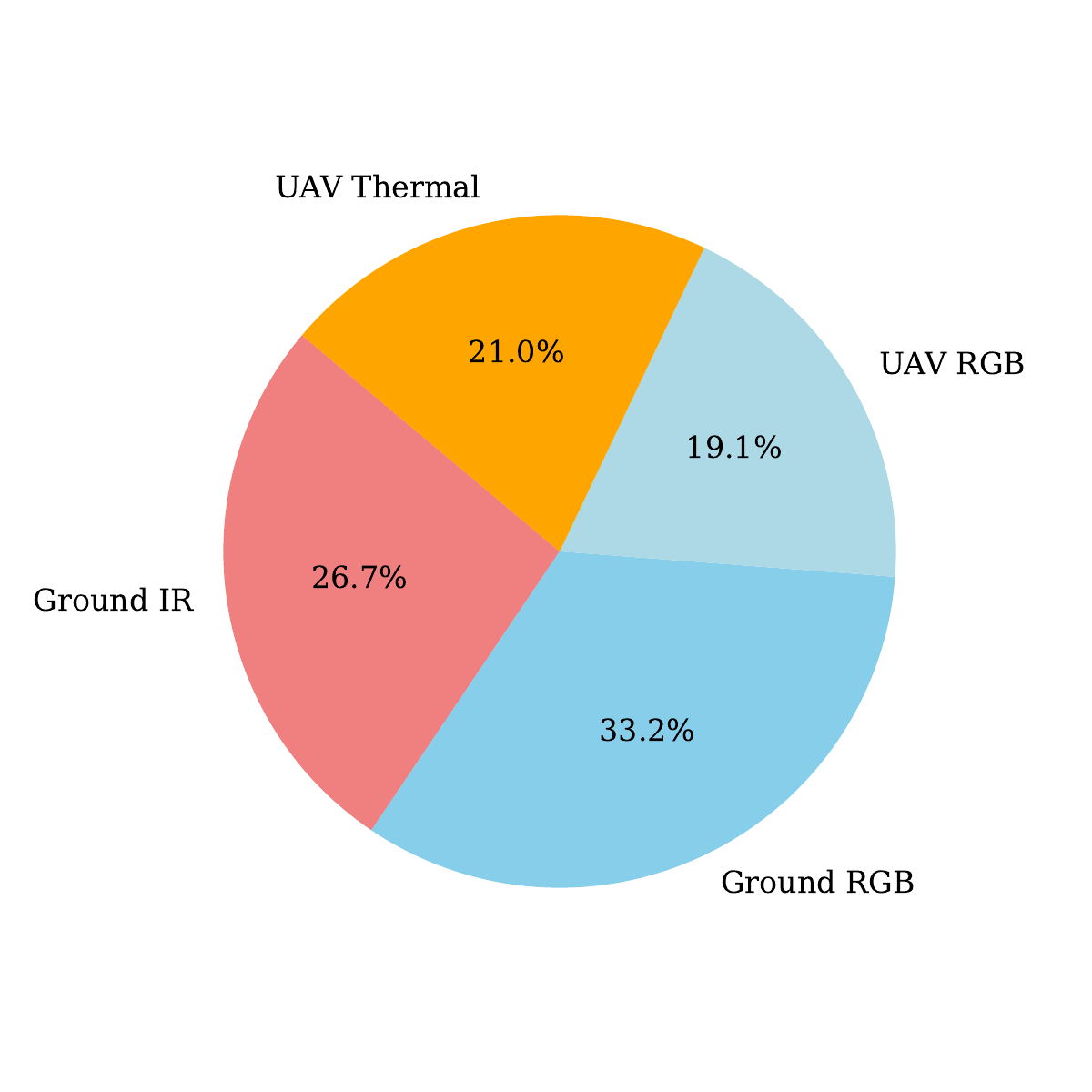}
    \caption{The proportion of bounding boxes in different modalities.}
    \label{fig:pie-b}
  \end{subfigure}
  \caption{\textbf{Bounding box analysis of the MP-ReID dataset.} (a) shows the image proportion in different scenes, and (b) shows the image proposition in different modalities.}
  \label{fig:pie}
\end{figure*}

\subsection{Data Statistics}
In our dataset, each ground RGB camera captures an average of 497 IDs and 7,545 bounding boxes, each ground infrared camera captures an average of 369 IDs and 6,050 bounding boxes, the UAV's RGB camera captures 341 IDs and 26,046 bounding boxes, and the UAV's thermal camera captures 474 IDs and 28,543 bounding boxes. There are 346 persons captured by only one camera and 381 persons with less than 10 bounding boxes, most person are captured by 2-8 cameras and have 10-60 bounding boxes.
Since outdoor scenes pose significant challenges, such as varying lighting conditions, occlusions, and high pedestrian densities, making them essential for robust ReID performance, it is reasonable to have a larger proportion of outdoor scene data than indoor scenes.

As shown in Fig. \ref{fig:pie-a}, Fig. \ref{fig:pie-b}, the bounding box ratios of persons in outdoor, indoor and aerial view are $51.2\%$, $8.8\%$ and $40.1\%$, the bounding box ratios of persons in RGB, infrared and thermal modality are $52.3\%$, $26.7\%$ and $21.0\%$.

\begin{figure*}[htbp]
  \centering
  \begin{subfigure}{0.45\linewidth}
    \includegraphics[width=\linewidth]{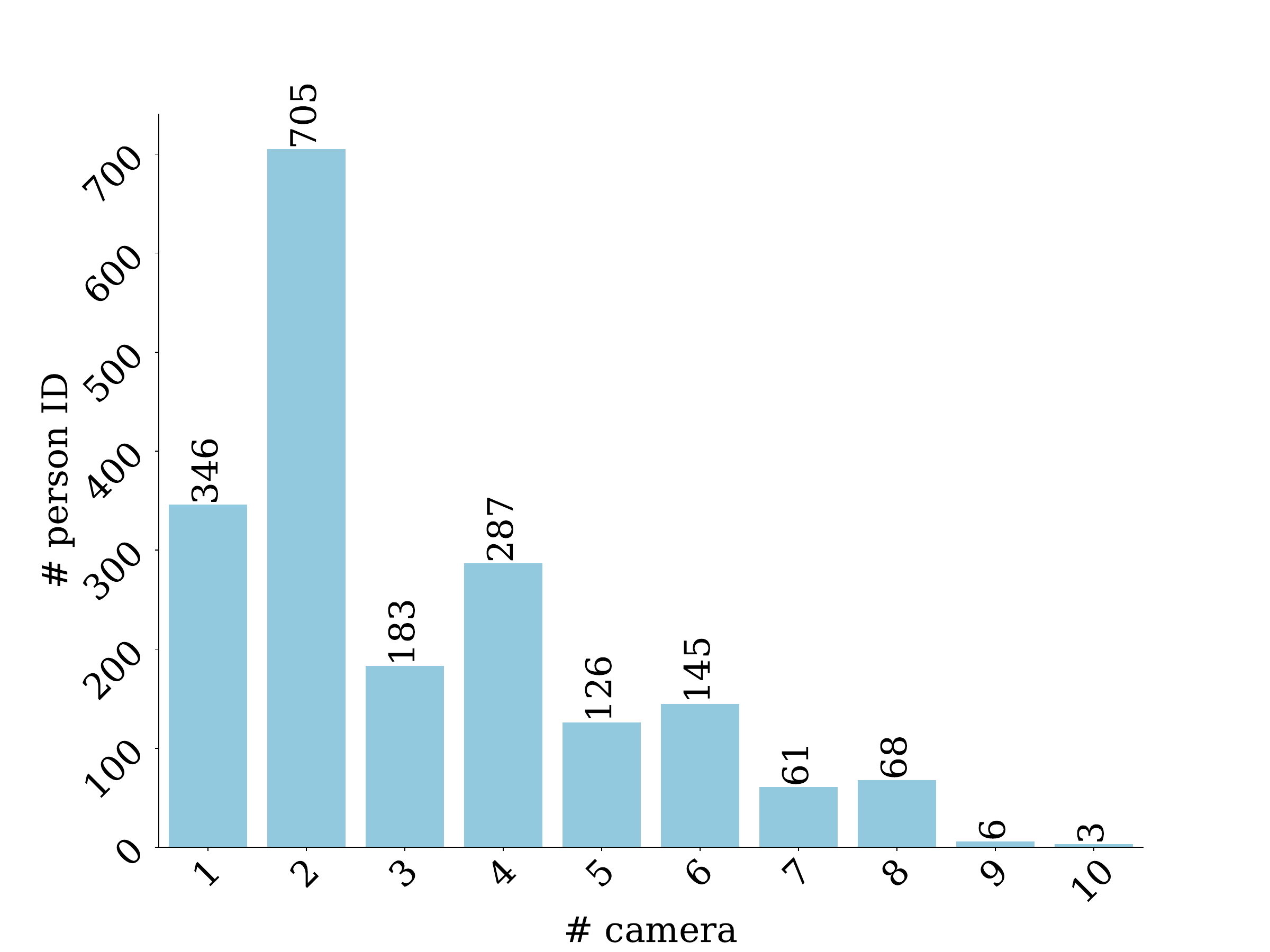}
    \caption{Distribution of person IDs Captured by different camera numbers.}
    \label{fig:camera-id}
  \end{subfigure} 
  \hfill 
  \begin{subfigure}{0.45\linewidth}
    \includegraphics[width=\linewidth]{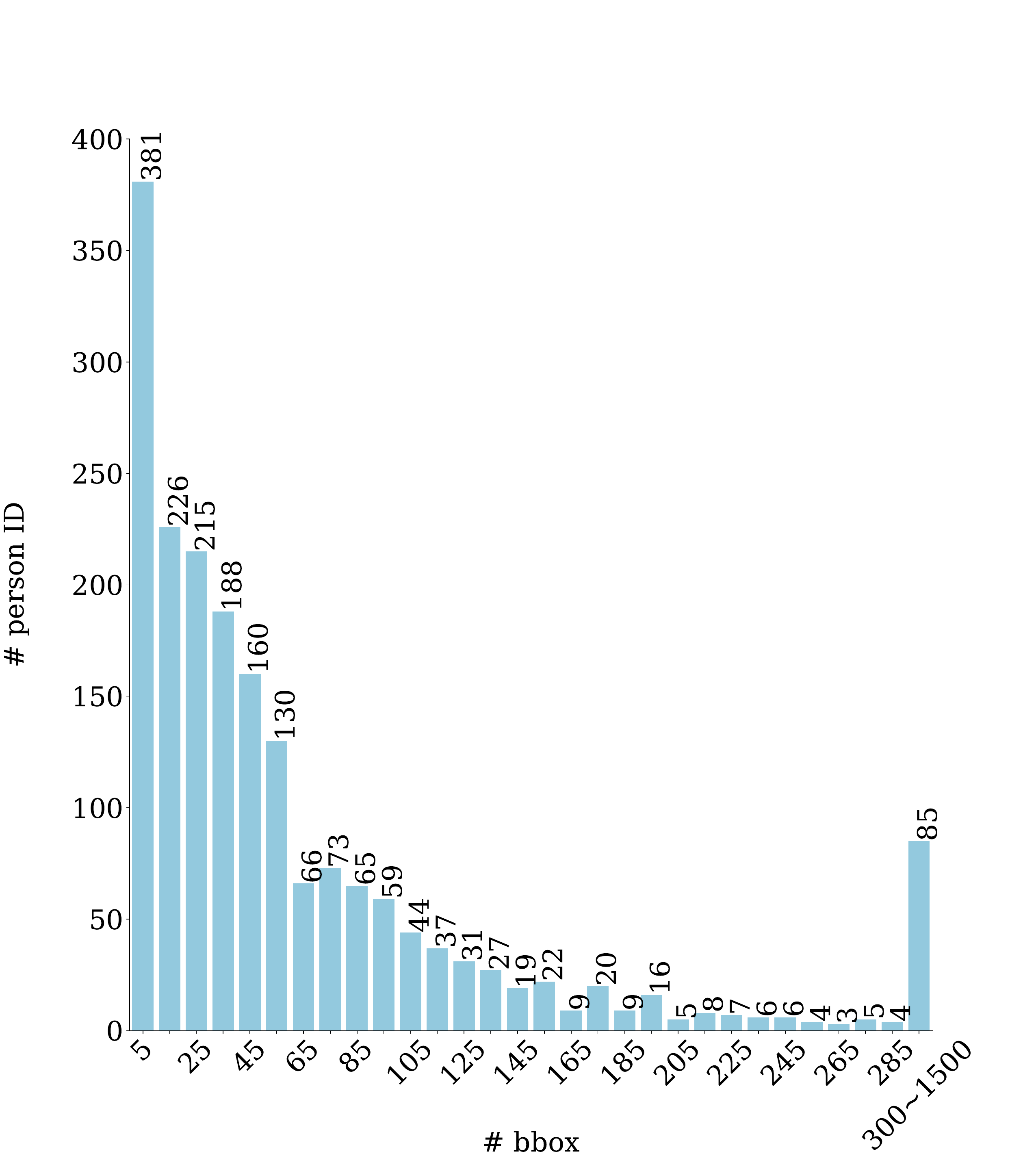}
    \caption{Distribution of person IDs across the number of bounding boxes.}
    \label{fig:camera-bb}
  \end{subfigure}
  \caption{Statistical analysis of the MP-ReID dataset.}
  \label{fig:stat}
\end{figure*}

Furthermore, we counted the number of cameras that captured the same pedestrian and the bounding box of each pedestrian. As shown in Fig. \ref{fig:camera-id} and Fig.\ref{fig:camera-bb}, there are over 870 persons captured by at least 3 cameras and only 346 persons appear once, which are considered as distractors in the later experiments. And the distribution of person IDs shows that most IDs in the dataset have over 25 bounding boxes, which is very enough to the ReID task. 

\begin{table*}[htbp]
\centering
\resizebox{\linewidth}{!}{
\begin{tabular}{cccccccccccc}
\toprule
\hline
\multicolumn{3}{c}{\multirow{2}{*}{Method}}   & \multicolumn{3}{c|}{$\text{U}_{T} \rightarrow \text{G}_{R}$} & \multicolumn{3}{c|}{$\text{U}_{T} \rightarrow \text{G}_{I}$}  & \multicolumn{3}{c}{$\text{U}_{R} \rightarrow \text{U}_{T}$}   \\
\multicolumn{3}{c}{}                                          & \multicolumn{1}{c}{Rank-1} & \multicolumn{1}{c}{Rank-5}  & \multicolumn{1}{c|}{mAP} & \multicolumn{1}{c}{Rank-1} & \multicolumn{1}{c}{Rank-5}  & \multicolumn{1}{c|}{mAP} & \multicolumn{1}{c}{Rank-1} & \multicolumn{1}{c}{Rank-5}  & \multicolumn{1}{c}{mAP} 
\\ \hline

\multicolumn{2}{l}{CAJ}               &  & \multicolumn{1}{c}{2.26}  & \multicolumn{1}{c}{10.32} & \multicolumn{1}{c|}{2.58}     & \multicolumn{1}{c}{1.69}  & \multicolumn{1}{c}{7.43} & \multicolumn{1}{c|}{1.93} & \multicolumn{1}{c}{22.14}  & \multicolumn{1}{c}{45.80} & \multicolumn{1}{c}{13.25}  \\ 

\multicolumn{2}{l}{\multirow{1}{*}{CAJ$_+$}}               &  & \multicolumn{1}{c}{9.68}  & \multicolumn{1}{c}{26.71} & \multicolumn{1}{c|}{7.20}     & \multicolumn{1}{c}{10.00}  & \multicolumn{1}{c}{25.51} & \multicolumn{1}{c|}{8.41} & \multicolumn{1}{c}{32.77}  & \multicolumn{1}{c}{61.07} & \multicolumn{1}{c}{21.34} \\ 

\multicolumn{2}{l}{\multirow{1}{*}{AGW}}               &  & \multicolumn{1}{c}{11.03}  & \multicolumn{1}{c}{25.55} & \multicolumn{1}{c|}{8.11}     & \multicolumn{1}{c}{11.10}  & \multicolumn{1}{c}{23.09} & \multicolumn{1}{c|}{8.34} & \multicolumn{1}{c}{24.20}  & \multicolumn{1}{c}{48.84} & \multicolumn{1}{c}{15.08} \\ 

\multicolumn{2}{l}{\multirow{1}{*}{DEEN}}               &  & \multicolumn{1}{c}{17.74}  & \multicolumn{1}{c}{40.84} & \multicolumn{1}{c|}{13.59}     & \multicolumn{1}{c}{16.62}  & \multicolumn{1}{c}{32.87} & \multicolumn{1}{c|}{11.00} & \multicolumn{1}{c}{56.61}  & \multicolumn{1}{c}{79.64} & \multicolumn{1}{c}{37.76}  \\ 

\multicolumn{2}{l}{\multirow{1}{*}{OTLA-ReID}}               &  & \multicolumn{1}{c}{21.29}  & \multicolumn{1}{c}{41.29} & \multicolumn{1}{c|}{12.89}     & \multicolumn{1}{c}{15.44}  & \multicolumn{1}{c}{36.76} & \multicolumn{1}{c|}{11.17} & \multicolumn{1}{c}{54.46}  & \multicolumn{1}{c}{75.00} & \multicolumn{1}{c}{35.72}\\ 

\multicolumn{2}{l}{\multirow{1}{*}{SAAI}}               &  & \multicolumn{1}{c}{18.93}  & \multicolumn{1}{c}{38.87} & \multicolumn{1}{c|}{12.76}     & \multicolumn{1}{c}{19.94}  & \multicolumn{1}{c}{41.27} & \multicolumn{1}{c|}{14.04} & \multicolumn{1}{c}{29.04}  & \multicolumn{1}{c}{62.07} & \multicolumn{1}{c}{21.49} \\ 
\hline
\multicolumn{2}{l}{\multirow{1}{*}{CSDN}}               &  & \multicolumn{1}{c}{10.42}  & \multicolumn{1}{c}{34.94} & \multicolumn{1}{c|}{11.66}     & \multicolumn{1}{c}{7.61}  & \multicolumn{1}{c}{10.01} & \multicolumn{1}{c|}{6.75} & \multicolumn{1}{c}{15.98}  & \multicolumn{1}{c}{25.30} & \multicolumn{1}{c}{8.97} \\ 
\hline

\multicolumn{3}{c}{\multirow{2}{*}{Method}}   & \multicolumn{3}{c|}{$\text{G}_{I} \rightarrow \text{U}_{R}$} & \multicolumn{3}{c|}{$\text{U}_{R} \rightarrow \text{G}_{R}$}  & \multicolumn{3}{c}{$\text{G}_{I} \rightarrow \text{G}_{R}$}   \\
\multicolumn{3}{c}{}                                          & \multicolumn{1}{c}{Rank-1} & \multicolumn{1}{c}{Rank-5}  & \multicolumn{1}{c|}{mAP} & \multicolumn{1}{c}{Rank-1} & \multicolumn{1}{c}{Rank-5}  & \multicolumn{1}{c|}{mAP} & \multicolumn{1}{c}{Rank-1} & \multicolumn{1}{c}{Rank-5}  & \multicolumn{1}{c}{mAP} 
\\ \hline
\multicolumn{2}{l}{CAJ}               &  & \multicolumn{1}{c}{23.50}  & \multicolumn{1}{c}{44.86} & \multicolumn{1}{c|}{16.88}     & \multicolumn{1}{c}{35.31}  & \multicolumn{1}{c}{56.99} & \multicolumn{1}{c|}{23.97} & \multicolumn{1}{c}{67.43}  & \multicolumn{1}{c}{82.25} & \multicolumn{1}{c}{37.36}  \\ 

\multicolumn{2}{l}{\multirow{1}{*}{CAJ$_+$}}               &  & \multicolumn{1}{c}{41.24}  & \multicolumn{1}{c}{61.98} & \multicolumn{1}{c|}{28.86}     & \multicolumn{1}{c}{40.62}  & \multicolumn{1}{c}{62.48} & \multicolumn{1}{c|}{32.76} & \multicolumn{1}{c}{83.07}  & \multicolumn{1}{c}{91.23} & \multicolumn{1}{c}{53.19} \\ 

\multicolumn{2}{l}{\multirow{1}{*}{AGW}}               &  & \multicolumn{1}{c}{33.11}  & \multicolumn{1}{c}{56.16} & \multicolumn{1}{c|}{25.95}     & \multicolumn{1}{c}{47.17}  & \multicolumn{1}{c}{69.12} & \multicolumn{1}{c|}{34.17} & \multicolumn{1}{c}{78.80}  & \multicolumn{1}{c}{90.23} & \multicolumn{1}{c}{51.24} \\ 

\multicolumn{2}{l}{\multirow{1}{*}{DEEN}}               &  & \multicolumn{1}{c}{44.75}  & \multicolumn{1}{c}{68.53} & \multicolumn{1}{c|}{30.74}     & \multicolumn{1}{c}{54.60}  & \multicolumn{1}{c}{75.22} & \multicolumn{1}{c|}{44.14} & \multicolumn{1}{c}{84.56}  & \multicolumn{1}{c}{92.57} & \multicolumn{1}{c}{57.40}  \\ 

\multicolumn{2}{l}{\multirow{1}{*}{OTLA-ReID}}               &  & \multicolumn{1}{c}{39.55}  & \multicolumn{1}{c}{70.06} & \multicolumn{1}{c|}{31.89}     & \multicolumn{1}{c}{69.91}  & \multicolumn{1}{c}{85.84} & \multicolumn{1}{c|}{44.14} & \multicolumn{1}{c}{82.65}  & \multicolumn{1}{c}{92.44} & \multicolumn{1}{c}{56.32}\\ 

\multicolumn{2}{l}{\multirow{1}{*}{SAAI}}               &  & \multicolumn{1}{c}{43.68}  & \multicolumn{1}{c}{66.03} & \multicolumn{1}{c|}{32.42}     & \multicolumn{1}{c}{65.88}  & \multicolumn{1}{c}{77.56} & \multicolumn{1}{c|}{51.86} & \multicolumn{1}{c}{84.82}  & \multicolumn{1}{c}{92.31} & \multicolumn{1}{c}{58.59} \\ 
\hline
\multicolumn{2}{l}{\multirow{1}{*}{CSDN}}               &  & \multicolumn{1}{c}{6.34}  & \multicolumn{1}{c}{17.22} & \multicolumn{1}{c|}{14.75}     & \multicolumn{1}{c}{27.54}  & \multicolumn{1}{c}{50.12} & \multicolumn{1}{c|}{19.66} & \multicolumn{1}{c}{68.72}  & \multicolumn{1}{c}{81.38} & \multicolumn{1}{c}{37.64}  \\ 
\hline
\bottomrule
\end{tabular}
}
\caption{The results of distractors added dataset for all baseline methods. Both rank accuracy ($\%$) and mAP($\%$) are reported.}
\label{tab:distractor results}
\end{table*}

\section{Experiment}
Furthermore, we categorized these 12 experimental settings into three classes: cross-modal only, cross-platform only, and both cross-modal and cross-platform. Specifically, cross-platform only: $\text{U}_{R} \rightarrow \text{G}_{R}$, $\text{G}_{R} \rightarrow \text{U}_{R}$; cross-modal only: $\text{G}_{I} \rightarrow \text{G}_{R}$, $\text{G}_{R} \rightarrow \text{G}_{I}$, $\text{U}_{T} \rightarrow \text{U}_{R}$, $\text{U}_{R} \rightarrow \text{U}_{T}$; cross-modal \& platform: $\text{U}_{R} \rightarrow \text{G}_{I}$, $\text{G}_{I} \rightarrow \text{U}_{R}$, $\text{U}_{T} \rightarrow \text{G}_{R}$, $\text{G}_{R} \rightarrow \text{U}_{T}$, $\text{U}_{T} \rightarrow \text{G}_{I}$, $\text{G}_{I} \rightarrow \text{U}_{T}$. We show the result of total 12 experimental settings in Table.~\ref{tab:results} in detail. What's more, the performance in each setting for the ablation study can be found in Table.~\ref{tab:ablationresults}.

\begin{table*}[htbp]
\centering
\resizebox{\linewidth}{!}{
\begin{tabular}{ccccccccccccccc}
\toprule
\hline
\multicolumn{3}{c}{\multirow{2}{*}{Method}}   & \multicolumn{3}{c|}{$\text{U}_{R} \rightarrow \text{G}_{R}$} & \multicolumn{3}{c|}{$\text{G}_{R} \rightarrow \text{U}_{R}$}  & \multicolumn{3}{c|}{$\text{G}_{I} \rightarrow \text{G}_{R}$} & \multicolumn{3}{c}{$\text{G}_{R} \rightarrow \text{G}_{I}$}   \\
\multicolumn{3}{c}{}                                          & \multicolumn{1}{c}{Rank-1} & \multicolumn{1}{c}{Rank-5}  & \multicolumn{1}{c|}{mAP} & \multicolumn{1}{c}{Rank-1} & \multicolumn{1}{c}{Rank-5}  & \multicolumn{1}{c|}{mAP} & \multicolumn{1}{c}{Rank-1} & \multicolumn{1}{c}{Rank-5}  & \multicolumn{1}{c|}{mAP} & \multicolumn{1}{c}{Rank-1} & \multicolumn{1}{c}{Rank-5}  & \multicolumn{1}{c}{mAP}
\\ \hline

\multicolumn{2}{l}{CAJ}               &  & \multicolumn{1}{c}{39.12 }  & \multicolumn{1}{c}{63.14 } & \multicolumn{1}{c|}{30.40 }     & \multicolumn{1}{c}{41.59 }  & \multicolumn{1}{c}{64.48 } & \multicolumn{1}{c|}{29.14 } & \multicolumn{1}{c}{68.68}  & \multicolumn{1}{c}{82.75} & \multicolumn{1}{c|}{38.19} & \multicolumn{1}{c}{68.32}  & \multicolumn{1}{c}{81.03} & \multicolumn{1}{c}{40.80}  \\ 

\multicolumn{2}{l}{\multirow{1}{*}{CAJ}$_+$}               &  & \multicolumn{1}{c}{41.47 }  & \multicolumn{1}{c}{66.57 } & \multicolumn{1}{c|}{36.14 }     & \multicolumn{1}{c}{53.72 }  & \multicolumn{1}{c}{73.60 } & \multicolumn{1}{c|}{38.92 } & \multicolumn{1}{c}{84.60}  & \multicolumn{1}{c}{91.57} & \multicolumn{1}{c|}{54.25} & \multicolumn{1}{c}{77.05}  & \multicolumn{1}{c}{87.38} & \multicolumn{1}{c}{53.16}  \\

\multicolumn{2}{l}{\multirow{1}{*}{AGW}}               &  & \multicolumn{1}{c}{48.53 }  & \multicolumn{1}{c}{71.96 } & \multicolumn{1}{c|}{41.12 }     & \multicolumn{1}{c}{58.83 }  & \multicolumn{1}{c}{79.79 } & \multicolumn{1}{c|}{44.63 } & \multicolumn{1}{c}{80.38}  & \multicolumn{1}{c}{90.75} & \multicolumn{1}{c|}{52.42} & \multicolumn{1}{c}{75.60}  & \multicolumn{1}{c}{86.90} & \multicolumn{1}{c}{51.92}  \\

\multicolumn{2}{l}{\multirow{1}{*}{DEEN}}               &  & \multicolumn{1}{c}{56.96 }  & \multicolumn{1}{c}{77.45 } & \multicolumn{1}{c|}{49.95 }     & \multicolumn{1}{c}{63.14 }  & \multicolumn{1}{c}{82.18 } & \multicolumn{1}{c|}{48.08 } & \multicolumn{1}{c}{85.89}  & \multicolumn{1}{c}{92.82} & \multicolumn{1}{c|}{57.57} & \multicolumn{1}{c}{79.05}  & \multicolumn{1}{c}{88.47} & \multicolumn{1}{c}{56.59}  \\

\multicolumn{2}{l}{\multirow{1}{*}{OTLA-ReID}}               &  & \multicolumn{1}{c}{74.51  }  & \multicolumn{1}{c}{83.33 } & \multicolumn{1}{c|}{64.55 }     & \multicolumn{1}{c}{71.97}  & \multicolumn{1}{c}{87.87 } & \multicolumn{1}{c|}{59.15 } & \multicolumn{1}{c}{84.13}  & \multicolumn{1}{c}{92.57} & \multicolumn{1}{c|}{57.49} & \multicolumn{1}{c}{78.51}  & \multicolumn{1}{c}{89.17} & \multicolumn{1}{c}{55.82}  \\

\multicolumn{2}{l}{\multirow{1}{*}{SAAI}}              &  & \multicolumn{1}{c}{68.87}  & \multicolumn{1}{c}{81.24} & \multicolumn{1}{c|}{53.88}     & \multicolumn{1}{c}{67.34}  & \multicolumn{1}{c}{84.17} & \multicolumn{1}{c|}{53.11} & \multicolumn{1}{c}{86.06}  & \multicolumn{1}{c}{92.73} & \multicolumn{1}{c|}{59.89} & \multicolumn{1}{c}{80.42}  & \multicolumn{1}{c}{89.99} & \multicolumn{1}{c}{57.84}  \\ 

\multicolumn{2}{l}{\multirow{1}{*}{CSDN}}              &  & \multicolumn{1}{c}{29.41}  & \multicolumn{1}{c}{53.92} & \multicolumn{1}{c|}{23.21}     & \multicolumn{1}{c}{33.89}  & \multicolumn{1}{c}{54.39} & \multicolumn{1}{c|}{23.23} & \multicolumn{1}{c}{71.34}  & \multicolumn{1}{c}{84.92} & \multicolumn{1}{c|}{42.86} & \multicolumn{1}{c}{76.17}  & \multicolumn{1}{c}{87.81} & \multicolumn{1}{c}{56.80}  \\

\multicolumn{2}{l}{\multirow{1}{*}{\textbf{Ours}}}              &  & \multicolumn{1}{c}{77.20}  & \multicolumn{1}{c}{87.62} & \multicolumn{1}{c|}{74.16}     & \multicolumn{1}{c}{80.34}  & \multicolumn{1}{c}{89.13} & \multicolumn{1}{c|}{74.41} & \multicolumn{1}{c}{86.25}  & \multicolumn{1}{c}{92.72} & \multicolumn{1}{c|}{71.21} & \multicolumn{1}{c}{81.77}  & \multicolumn{1}{c}{89.45} & \multicolumn{1}{c}{69.30}  \\

\toprule

\hline
\multicolumn{3}{c}{\multirow{2}{*}{}}   & \multicolumn{3}{c|}{$\text{U}_{T} \rightarrow \text{U}_{R}$} & \multicolumn{3}{c|}{$\text{U}_{R} \rightarrow \text{U}_{T}$}  & \multicolumn{3}{c|}{$\text{U}_{R} \rightarrow \text{G}_{I}$} & \multicolumn{3}{c}{$\text{G}_{I} \rightarrow \text{U}_{R}$}   \\
\multicolumn{3}{c}{}                                          & \multicolumn{1}{c}{Rank-1} & \multicolumn{1}{c}{Rank-5}  & \multicolumn{1}{c|}{mAP} & \multicolumn{1}{c}{Rank-1} & \multicolumn{1}{c}{Rank-5}  & \multicolumn{1}{c|}{mAP} & \multicolumn{1}{c}{Rank-1} & \multicolumn{1}{c}{Rank-5}  & \multicolumn{1}{c|}{mAP} & \multicolumn{1}{c}{Rank-1} & \multicolumn{1}{c}{Rank-5}  & \multicolumn{1}{c}{mAP}
\\ \hline

\multicolumn{2}{l}{CAJ~}               &  & \multicolumn{1}{c}{20.62 }  & \multicolumn{1}{c}{46.25 } & \multicolumn{1}{c|}{14.30 }     & \multicolumn{1}{c}{23.75 }  & \multicolumn{1}{c}{48.75 } & \multicolumn{1}{c|}{14.08 } & \multicolumn{1}{c}{28.88 }  & \multicolumn{1}{c}{57.08 } & \multicolumn{1}{c|}{19.65 } & \multicolumn{1}{c}{24.92 }  & \multicolumn{1}{c}{46.67 } & \multicolumn{1}{c}{17.57 }  \\ 

\multicolumn{2}{l}{\multirow{1}{*}{CAJ$_+$}}               &  & \multicolumn{1}{c}{36.61 }  & \multicolumn{1}{c}{62.32 } & \multicolumn{1}{c|}{24.56 }     & \multicolumn{1}{c}{34.38 }  & \multicolumn{1}{c}{63.39 } & \multicolumn{1}{c|}{22.10 } & \multicolumn{1}{c}{43.60 }  & \multicolumn{1}{c}{71.01 } & \multicolumn{1}{c|}{32.48 } & \multicolumn{1}{c}{43.67 }  & \multicolumn{1}{c}{64.97 } & \multicolumn{1}{c}{30.42 }  \\

\multicolumn{2}{l}{\multirow{1}{*}{AGW}}               &  & \multicolumn{1}{c}{25.18 }  & \multicolumn{1}{c}{48.30 } & \multicolumn{1}{c|}{15.25 }     & \multicolumn{1}{c}{26.34 }  & \multicolumn{1}{c}{51.34 } & \multicolumn{1}{c|}{16.00 } & \multicolumn{1}{c}{35.73 }  & \multicolumn{1}{c}{64.16 } & \multicolumn{1}{c|}{28.39 } & \multicolumn{1}{c}{34.80 }  & \multicolumn{1}{c}{57.47 } & \multicolumn{1}{c}{27.42 }  \\

\multicolumn{2}{l}{\multirow{1}{*}{DEEN}}               &  & \multicolumn{1}{c}{55.36}  & \multicolumn{1}{c}{78.57} & \multicolumn{1}{c|}{40.55}     & \multicolumn{1}{c}{58.04}  & \multicolumn{1}{c}{82.32} & \multicolumn{1}{c|}{40.79} & \multicolumn{1}{c}{46.72 }  & \multicolumn{1}{c}{71.36 } & \multicolumn{1}{c|}{32.20 } & \multicolumn{1}{c}{46.85 }  & \multicolumn{1}{c}{75.06 } & \multicolumn{1}{c}{37.64 }  \\ 

\multicolumn{2}{l}{\multirow{1}{*}{OTLA-ReID}}               &  & \multicolumn{1}{c}{53.57 }  & \multicolumn{1}{c}{81.25 } & \multicolumn{1}{c|}{35.87 }     & \multicolumn{1}{c}{56.25 }  & \multicolumn{1}{c}{75.89 } & \multicolumn{1}{c|}{37.08 } & \multicolumn{1}{c}{56.18 }  & \multicolumn{1}{c}{76.40 } & \multicolumn{1}{c|}{39.86 } & \multicolumn{1}{c}{41.81 }  & \multicolumn{1}{c}{72.88  } & \multicolumn{1}{c}{33.34  }  \\ 

\multicolumn{2}{l}{\multirow{1}{*}{SAAI}}               &  & \multicolumn{1}{c}{37.16}  & \multicolumn{1}{c}{63.52} & \multicolumn{1}{c|}{24.88}     & \multicolumn{1}{c}{38.03}  & \multicolumn{1}{c}{65.66} & \multicolumn{1}{c|}{24.21} & \multicolumn{1}{c}{47.74}  & \multicolumn{1}{c}{72.31} & \multicolumn{1}{c|}{32.53} & \multicolumn{1}{c}{43.68}  & \multicolumn{1}{c}{68.06} & \multicolumn{1}{c}{32.95}  \\ 

\multicolumn{2}{l}{\multirow{1}{*}{CSDN}}              &  & \multicolumn{1}{c}{10.71}  & \multicolumn{1}{c}{24.11} & \multicolumn{1}{c|}{15.43}     & \multicolumn{1}{c}{7.14}  & \multicolumn{1}{c}{18.76} & \multicolumn{1}{c|}{16.14} & \multicolumn{1}{c}{24.72}  & \multicolumn{1}{c}{44.45} & \multicolumn{1}{c|}{13.44} & \multicolumn{1}{c}{16.95}  & \multicolumn{1}{c}{28.25} & \multicolumn{1}{c}{10.10}  \\

\multicolumn{2}{l}{\multirow{1}{*}{\textbf{Ours}}}              &  & \multicolumn{1}{c}{56.32}  & \multicolumn{1}{c}{80.33} & \multicolumn{1}{c|}{52.46}     & \multicolumn{1}{c}{58.70}  & \multicolumn{1}{c}{77.51} & \multicolumn{1}{c|}{50.26} & \multicolumn{1}{c}{61.28}  & \multicolumn{1}{c}{79.24} & \multicolumn{1}{c|}{45.88} & \multicolumn{1}{c}{61.67}  & \multicolumn{1}{c}{78.12} & \multicolumn{1}{c}{49.77}  \\ 

\toprule
\hline
\multicolumn{3}{c}{\multirow{2}{*}{}}   & \multicolumn{3}{c|}{$\text{U}_{T} \rightarrow \text{G}_{R}$} & \multicolumn{3}{c|}{$\text{G}_{R} \rightarrow \text{U}_{T}$}  & \multicolumn{3}{c|}{$\text{U}_{T} \rightarrow \text{G}_{I}$} & \multicolumn{3}{c}{$\text{G}_{I} \rightarrow \text{U}_{T}$}   \\
\multicolumn{3}{c}{}                                          & \multicolumn{1}{c}{Rank-1} & \multicolumn{1}{c}{Rank-5}  & \multicolumn{1}{c|}{mAP} & \multicolumn{1}{c}{Rank-1} & \multicolumn{1}{c}{Rank-5}  & \multicolumn{1}{c|}{mAP} & \multicolumn{1}{c}{Rank-1} & \multicolumn{1}{c}{Rank-5}  & \multicolumn{1}{c|}{mAP} & \multicolumn{1}{c}{Rank-1} & \multicolumn{1}{c}{Rank-5}  & \multicolumn{1}{c}{mAP}
\\ \hline

\multicolumn{2}{l}{CAJ}               &  & \multicolumn{1}{c}{2.58 }  & \multicolumn{1}{c}{11.81 } & \multicolumn{1}{c|}{2.84  }     & \multicolumn{1}{c}{3.13 }  & \multicolumn{1}{c}{11.91 } & \multicolumn{1}{c|}{2.96} & \multicolumn{1}{c}{2.13 }  & \multicolumn{1}{c}{8.46 } & \multicolumn{1}{c|}{1.87} & \multicolumn{1}{c}{2.09}  & \multicolumn{1}{c}{9.66} & \multicolumn{1}{c}{2.55}  \\ 

\multicolumn{2}{l}{\multirow{1}{*}{CAJ$_+$}}               &  & \multicolumn{1}{c}{10.71}  & \multicolumn{1}{c}{29.35} & \multicolumn{1}{c|}{7.78}     & \multicolumn{1}{c}{8.42}  & \multicolumn{1}{c}{25.93} & \multicolumn{1}{c|}{6.39} & \multicolumn{1}{c}{12.06}  & \multicolumn{1}{c}{28.97} & \multicolumn{1}{c|}{9.12} & \multicolumn{1}{c}{10.6}  & \multicolumn{1}{c}{29.45} & \multicolumn{1}{c}{8.56}  \\

\multicolumn{2}{l}{\multirow{1}{*}{AGW}}               &  & \multicolumn{1}{c}{11.74}  & \multicolumn{1}{c}{27.61} & \multicolumn{1}{c|}{8.61}     & \multicolumn{1}{c}{8.97}  & \multicolumn{1}{c}{26.11} & \multicolumn{1}{c|}{7.83} & \multicolumn{1}{c}{13.16}  & \multicolumn{1}{c}{25.66} & \multicolumn{1}{c|}{9.11} & \multicolumn{1}{c}{10.85}  & \multicolumn{1}{c}{26.64} & \multicolumn{1}{c}{8.29}  \\

\multicolumn{2}{l}{\multirow{1}{*}{DEEN}}               &  & \multicolumn{1}{c}{20.19}  & \multicolumn{1}{c}{45.61} & \multicolumn{1}{c|}{14.12}     & \multicolumn{1}{c}{18.92}  & \multicolumn{1}{c}{43.4} & \multicolumn{1}{c|}{14.25} & \multicolumn{1}{c}{17.65}  & \multicolumn{1}{c}{36.47} & \multicolumn{1}{c|}{11.78} & \multicolumn{1}{c}{15.23}  & \multicolumn{1}{c}{38.64} & \multicolumn{1}{c}{10.40}  \\ 

\multicolumn{2}{l}{\multirow{1}{*}{OTLA-ReID}}               &  & \multicolumn{1}{c}{23.87 }  & \multicolumn{1}{c}{42.58 } & \multicolumn{1}{c|}{13.62 }     & \multicolumn{1}{c}{17.33  }  & \multicolumn{1}{c}{41.64 } & \multicolumn{1}{c|}{12.29 } & \multicolumn{1}{c}{18.38 }  & \multicolumn{1}{c}{39.71 } & \multicolumn{1}{c|}{11.75 } & \multicolumn{1}{c}{18.30 }  & \multicolumn{1}{c}{46.38 } & \multicolumn{1}{c}{13.08 }  \\

\multicolumn{2}{l}{\multirow{1}{*}{SAAI}}               &  & \multicolumn{1}{c}{20.39}  & \multicolumn{1}{c}{42.41} & \multicolumn{1}{c|}{13.47}     & \multicolumn{1}{c}{19.55}  & \multicolumn{1}{c}{42.33} & \multicolumn{1}{c|}{15.31} & \multicolumn{1}{c}{20.25}  & \multicolumn{1}{c}{43.30} & \multicolumn{1}{c|}{14.55} & \multicolumn{1}{c}{18.77}  & \multicolumn{1}{c}{41.14} & \multicolumn{1}{c}{13.04}  \\ 

\multicolumn{2}{l}{\multirow{1}{*}{CSDN}}              &  & \multicolumn{1}{c}{11.58}  & \multicolumn{1}{c}{16.10} & \multicolumn{1}{c|}{10.62}     & \multicolumn{1}{c}{8.82}  & \multicolumn{1}{c}{12.48} & \multicolumn{1}{c|}{8.47} & \multicolumn{1}{c}{8.94}  & \multicolumn{1}{c}{11.15} & \multicolumn{1}{c|}{7.39} & \multicolumn{1}{c}{7.98}  & \multicolumn{1}{c}{10.53} & \multicolumn{1}{c}{6.38}  \\

\multicolumn{2}{l}{\multirow{1}{*}{\textbf{Ours}}}              &  & \multicolumn{1}{c}{34.71}  & \multicolumn{1}{c}{50.48} & \multicolumn{1}{c|}{34.14}     & \multicolumn{1}{c}{31.82}  & \multicolumn{1}{c}{48.16} & \multicolumn{1}{c|}{37.23} & \multicolumn{1}{c}{35.39}  & \multicolumn{1}{c}{50.21} & \multicolumn{1}{c|}{34.65} & \multicolumn{1}{c}{34.08}  & \multicolumn{1}{c}{57.39} & \multicolumn{1}{c}{39.89}  \\ 

\hline
\bottomrule
\end{tabular}
}
\caption{The results of all baseline methods. Both rank accuracy (\%) and mAP(\%) are reported. $\text{U}_T$, $\text{U}_R$, $\text{G}_I$ and $\text{G}_R$ stand for UAV thermal, UAV RGB, ground infrared and ground RGB, respectively.}
\label{tab:results}
\end{table*}

\begin{table*}[htbp]
\centering
\resizebox{\linewidth}{!}{
\begin{tabular}{ccccccccccccccc}
\toprule
\hline
\multicolumn{3}{c}{\multirow{2}{*}{Method}}   & \multicolumn{3}{c|}{$\text{U}_{R} \rightarrow \text{G}_{R}$} & \multicolumn{3}{c|}{$\text{G}_{R} \rightarrow \text{U}_{R}$}  & \multicolumn{3}{c|}{$\text{G}_{I} \rightarrow \text{G}_{R}$} & \multicolumn{3}{c}{$\text{G}_{R} \rightarrow \text{G}_{I}$}   \\
\multicolumn{3}{c}{}                                          & \multicolumn{1}{c}{Rank-1} & \multicolumn{1}{c}{Rank-5}  & \multicolumn{1}{c|}{mAP} & \multicolumn{1}{c}{Rank-1} & \multicolumn{1}{c}{Rank-5}  & \multicolumn{1}{c|}{mAP} & \multicolumn{1}{c}{Rank-1} & \multicolumn{1}{c}{Rank-5}  & \multicolumn{1}{c|}{mAP} & \multicolumn{1}{c}{Rank-1} & \multicolumn{1}{c}{Rank-5}  & \multicolumn{1}{c}{mAP}
\\ \hline

\multicolumn{2}{l}{Base}               &  & \multicolumn{1}{c}{74.52 }  & \multicolumn{1}{c}{80.79 } & \multicolumn{1}{c|}{71.04 }     & \multicolumn{1}{c}{79.50 }  & \multicolumn{1}{c}{82.46 } & \multicolumn{1}{c|}{71.33 } & \multicolumn{1}{c}{85.55}  & \multicolumn{1}{c}{90.55} & \multicolumn{1}{c|}{69.01} & \multicolumn{1}{c}{82.88}  & \multicolumn{1}{c}{88.67} & \multicolumn{1}{c}{67.20}  \\ 

\multicolumn{2}{l}{\multirow{1}{*}{Base+MS Prompt}}               &  & \multicolumn{1}{c}{74.37 }  & \multicolumn{1}{c}{80.71 } & \multicolumn{1}{c|}{71.36 }     & \multicolumn{1}{c}{79.98 }  & \multicolumn{1}{c}{87.81 } & \multicolumn{1}{c|}{71.31 } & \multicolumn{1}{c}{85.62}  & \multicolumn{1}{c}{91.16} & \multicolumn{1}{c|}{70.51} & \multicolumn{1}{c}{83.94}  & \multicolumn{1}{c}{88.96} & \multicolumn{1}{c}{70.59}  \\

\multicolumn{2}{l}{\multirow{1}{*}{Base+PM Prompt} }              &  & \multicolumn{1}{c}{77.26 }  & \multicolumn{1}{c}{87.20 } & \multicolumn{1}{c|}{73.34 }     & \multicolumn{1}{c}{79.97}  & \multicolumn{1}{c}{88.81 } & \multicolumn{1}{c|}{74.21 } & \multicolumn{1}{c}{86.03}  & \multicolumn{1}{c}{91.66} & \multicolumn{1}{c|}{70.78} & \multicolumn{1}{c}{84.70}  & \multicolumn{1}{c}{89.06} & \multicolumn{1}{c}{70.51}  \\

\multicolumn{2}{l}{\multirow{1}{*}{Base+IE}}               &  & \multicolumn{1}{c}{74.77 }  & \multicolumn{1}{c}{81.21 } & \multicolumn{1}{c|}{71.22}     & \multicolumn{1}{c}{79.87 }  & \multicolumn{1}{c}{86.74 } & \multicolumn{1}{c|}{71.31 } & \multicolumn{1}{c}{85.62}  & \multicolumn{1}{c}{90.77} & \multicolumn{1}{c|}{70.38} & \multicolumn{1}{c}{83.03}  & \multicolumn{1}{c}{89.01} & \multicolumn{1}{c}{69.21}  \\

\multicolumn{2}{l}{\multirow{1}{*}{Base+MS Prompt+IE} }              &  & \multicolumn{1}{c}{74.81  }  & \multicolumn{1}{c}{84.21 } & \multicolumn{1}{c|}{72.95 }     & \multicolumn{1}{c}{80.10}  & \multicolumn{1}{c}{88.66 } & \multicolumn{1}{c|}{72.67 } & \multicolumn{1}{c}{85.88}  & \multicolumn{1}{c}{92.19} & \multicolumn{1}{c|}{71.29} & \multicolumn{1}{c}{84.12}  & \multicolumn{1}{c}{89.41} & \multicolumn{1}{c}{71.24}  \\

\multicolumn{2}{l}{\multirow{1}{*}{Base+PM Prompt+IE}}              &  & \multicolumn{1}{c}{77.20}  & \multicolumn{1}{c}{87.62} & \multicolumn{1}{c|}{74.16}     & \multicolumn{1}{c}{80.34}  & \multicolumn{1}{c}{89.13} & \multicolumn{1}{c|}{74.41} & \multicolumn{1}{c}{86.25}  & \multicolumn{1}{c}{92.72} & \multicolumn{1}{c|}{71.24} & \multicolumn{1}{c}{87.77}  & \multicolumn{1}{c}{89.45} & \multicolumn{1}{c}{69.30}  \\

\toprule

\hline
\multicolumn{3}{c}{\multirow{2}{*}{}}   & \multicolumn{3}{c|}{$\text{U}_{T} \rightarrow \text{U}_{R}$} & \multicolumn{3}{c|}{$\text{U}_{R} \rightarrow \text{U}_{T}$}  & \multicolumn{3}{c|}{$\text{U}_{R} \rightarrow \text{G}_{I}$} & \multicolumn{3}{c}{$\text{G}_{I} \rightarrow \text{U}_{R}$}   \\
\multicolumn{3}{c}{}                                          & \multicolumn{1}{c}{Rank-1} & \multicolumn{1}{c}{Rank-5}  & \multicolumn{1}{c|}{mAP} & \multicolumn{1}{c}{Rank-1} & \multicolumn{1}{c}{Rank-5}  & \multicolumn{1}{c|}{mAP} & \multicolumn{1}{c}{Rank-1} & \multicolumn{1}{c}{Rank-5}  & \multicolumn{1}{c|}{mAP} & \multicolumn{1}{c}{Rank-1} & \multicolumn{1}{c}{Rank-5}  & \multicolumn{1}{c}{mAP}
\\ \hline

\multicolumn{2}{l}{Base}               &  & \multicolumn{1}{c}{34.11}  & \multicolumn{1}{c}{45.34 } & \multicolumn{1}{c|}{31.40 }     & \multicolumn{1}{c}{41.88 }  & \multicolumn{1}{c}{58.27 } & \multicolumn{1}{c|}{31.11 } & \multicolumn{1}{c}{43.40 }  & \multicolumn{1}{c}{54.44} & \multicolumn{1}{c|}{35.41 } & \multicolumn{1}{c}{50.21 }  & \multicolumn{1}{c}{58.48 } & \multicolumn{1}{c}{41.87 }  \\ 

\multicolumn{2}{l}{\multirow{1}{*}{Base+MS Prompt}}               &  & \multicolumn{1}{c}{43.71 }  & \multicolumn{1}{c}{72.23 } & \multicolumn{1}{c|}{42.91 }     & \multicolumn{1}{c}{56.10 }  & \multicolumn{1}{c}{72.62 } & \multicolumn{1}{c|}{43.07 } & \multicolumn{1}{c}{50.51 }  & \multicolumn{1}{c}{63.91 } & \multicolumn{1}{c|}{40.62 } & \multicolumn{1}{c}{51.94 }  & \multicolumn{1}{c}{65.90 } & \multicolumn{1}{c}{43.51 }  \\

\multicolumn{2}{l}{\multirow{1}{*}{Base+PM Prompt} }              &  & \multicolumn{1}{c}{54.76 }  & \multicolumn{1}{c}{74.31 } & \multicolumn{1}{c|}{51.49 }     & \multicolumn{1}{c}{55.76 }  & \multicolumn{1}{c}{72.49 } & \multicolumn{1}{c|}{47.98 } & \multicolumn{1}{c}{59.52 }  & \multicolumn{1}{c}{72.64 } & \multicolumn{1}{c|}{41.59 } & \multicolumn{1}{c}{59.65 }  & \multicolumn{1}{c}{74.21 } & \multicolumn{1}{c}{49.68 }  \\

\multicolumn{2}{l}{\multirow{1}{*}{Base+IE}}               &  & \multicolumn{1}{c}{40.19}  & \multicolumn{1}{c}{66.35} & \multicolumn{1}{c|}{38.28}     & \multicolumn{1}{c}{47.17}  & \multicolumn{1}{c}{63.69} & \multicolumn{1}{c|}{42.73} & \multicolumn{1}{c}{48.57}  & \multicolumn{1}{c}{61.97 } & \multicolumn{1}{c|}{40.26 } & \multicolumn{1}{c}{51.46 }  & \multicolumn{1}{c}{63.89 } & \multicolumn{1}{c}{42.67 }  \\ 

\multicolumn{2}{l}{\multirow{1}{*}{Base+MS Prompt+IE}}               &  & \multicolumn{1}{c}{47.67 }  & \multicolumn{1}{c}{74.50 } & \multicolumn{1}{c|}{44.63 }     & \multicolumn{1}{c}{56.96}  & \multicolumn{1}{c}{74.80 } & \multicolumn{1}{c|}{47.48 } & \multicolumn{1}{c}{52.02 }  & \multicolumn{1}{c}{67.65 } & \multicolumn{1}{c|}{45.91 } & \multicolumn{1}{c}{54.61 }  & \multicolumn{1}{c}{70.33  } & \multicolumn{1}{c}{44.89  }  \\ 

\multicolumn{2}{l}{\multirow{1}{*}{Base+PM Prompt+IE}}               &  & \multicolumn{1}{c}{56.32}  & \multicolumn{1}{c}{80.33} & \multicolumn{1}{c|}{52.46}     & \multicolumn{1}{c}{58.70}  & \multicolumn{1}{c}{77.51} & \multicolumn{1}{c|}{50.26} & \multicolumn{1}{c}{61.28}  & \multicolumn{1}{c}{79.24} & \multicolumn{1}{c|}{45.88} & \multicolumn{1}{c}{61.67}  & \multicolumn{1}{c}{78.12} & \multicolumn{1}{c}{49.77}  \\

\toprule
\hline
\multicolumn{3}{c}{\multirow{2}{*}{}}   & \multicolumn{3}{c|}{$\text{U}_{T} \rightarrow \text{G}_{R}$} & \multicolumn{3}{c|}{$\text{G}_{R} \rightarrow \text{U}_{T}$}  & \multicolumn{3}{c|}{$\text{U}_{T} \rightarrow \text{G}_{I}$} & \multicolumn{3}{c}{$\text{G}_{I} \rightarrow \text{U}_{T}$}   \\
\multicolumn{3}{c}{}                                          & \multicolumn{1}{c}{Rank-1} & \multicolumn{1}{c}{Rank-5}  & \multicolumn{1}{c|}{mAP} & \multicolumn{1}{c}{Rank-1} & \multicolumn{1}{c}{Rank-5}  & \multicolumn{1}{c|}{mAP} & \multicolumn{1}{c}{Rank-1} & \multicolumn{1}{c}{Rank-5}  & \multicolumn{1}{c|}{mAP} & \multicolumn{1}{c}{Rank-1} & \multicolumn{1}{c}{Rank-5}  & \multicolumn{1}{c}{mAP}
\\ \hline

\multicolumn{2}{l}{Base}               &  & \multicolumn{1}{c}{21.04 }  & \multicolumn{1}{c}{28.53 } & \multicolumn{1}{c|}{17.81 }     & \multicolumn{1}{c}{19.61 }  & \multicolumn{1}{c}{30.02 } & \multicolumn{1}{c|}{14.87} & \multicolumn{1}{c}{17.14 }  & \multicolumn{1}{c}{28.09} & \multicolumn{1}{c|}{14.51} & \multicolumn{1}{c}{19.02}  & \multicolumn{1}{c}{31.64} & \multicolumn{1}{c}{14.03}  \\ 

\multicolumn{2}{l}{\multirow{1}{*}{Base+MS Prompt}}               &  & \multicolumn{1}{c}{23.67}  & \multicolumn{1}{c}{37.11} & \multicolumn{1}{c|}{22.51}     & \multicolumn{1}{c}{19.87}  & \multicolumn{1}{c}{34.39} & \multicolumn{1}{c|}{18.84} & \multicolumn{1}{c}{22.49}  & \multicolumn{1}{c}{32.52} & \multicolumn{1}{c|}{16.67} & \multicolumn{1}{c}{20.97}  & \multicolumn{1}{c}{43.23} & \multicolumn{1}{c}{19.34}  \\

\multicolumn{2}{l}{\multirow{1}{*}{Base+PM Prompt} }              &  & \multicolumn{1}{c}{32.86}  & \multicolumn{1}{c}{45.09} & \multicolumn{1}{c|}{30.87}     & \multicolumn{1}{c}{30.23}  & \multicolumn{1}{c}{44.84} & \multicolumn{1}{c|}{36.29} & \multicolumn{1}{c}{31.25}  & \multicolumn{1}{c}{47.93} & \multicolumn{1}{c|}{34.46} & \multicolumn{1}{c}{30.45}  & \multicolumn{1}{c}{54.59} & \multicolumn{1}{c}{38.51}  \\

\multicolumn{2}{l}{\multirow{1}{*}{Base+IE} }              &  & \multicolumn{1}{c}{23.16}  & \multicolumn{1}{c}{36.57} & \multicolumn{1}{c|}{20.23}     & \multicolumn{1}{c}{19.89}  & \multicolumn{1}{c}{34.01} & \multicolumn{1}{c|}{16.67} & \multicolumn{1}{c}{17.52}  & \multicolumn{1}{c}{30.08} & \multicolumn{1}{c|}{16.64} & \multicolumn{1}{c}{20.49}  & \multicolumn{1}{c}{35.57} & \multicolumn{1}{c}{19.28}  \\ 

\multicolumn{2}{l}{\multirow{1}{*}{Base+MS Prompt+IE} }              &  & \multicolumn{1}{c}{24.89 }  & \multicolumn{1}{c}{41.47 } & \multicolumn{1}{c|}{30.02 }     & \multicolumn{1}{c}{29.21  }  & \multicolumn{1}{c}{41.76 } & \multicolumn{1}{c|}{26.24} & \multicolumn{1}{c}{27.13 }  & \multicolumn{1}{c}{37.24} & \multicolumn{1}{c|}{28.88 } & \multicolumn{1}{c}{24.41 }  & \multicolumn{1}{c}{46.58 } & \multicolumn{1}{c}{28.80 }  \\

\multicolumn{2}{l}{\multirow{1}{*}{Base+PM Prompt+IE}}               &  & \multicolumn{1}{c}{34.71}  & \multicolumn{1}{c}{50.48} & \multicolumn{1}{c|}{34.14}     & \multicolumn{1}{c}{31.82}  & \multicolumn{1}{c}{48.16} & \multicolumn{1}{c|}{37.23} & \multicolumn{1}{c}{35.39}  & \multicolumn{1}{c}{50.21} & \multicolumn{1}{c|}{34.65} & \multicolumn{1}{c}{34.08}  & \multicolumn{1}{c}{57.39} & \multicolumn{1}{c}{39.89}  \\

\hline
\bottomrule
\end{tabular}
}
\caption{The results of all Ablation studys. Both rank accuracy (\%) and mAP(\%) are reported. $\text{U}_T$, $\text{U}_R$, $\text{G}_I$ and $\text{G}_R$ stand for UAV thermal, UAV RGB, ground infrared and ground RGB, respectively.}
\label{tab:ablationresults}
\end{table*}

\section{License}

The MP-ReID Dataset will be freely available, under the Creative Commons
Attribution-NonCommercial-NoDerivs 4.0 (CC BY-NC-ND 4.0) International license. 


\section{Visualization of Retrieval Results}

\begin{figure*}[htbp]
    \centering
    \includegraphics[width=\linewidth]{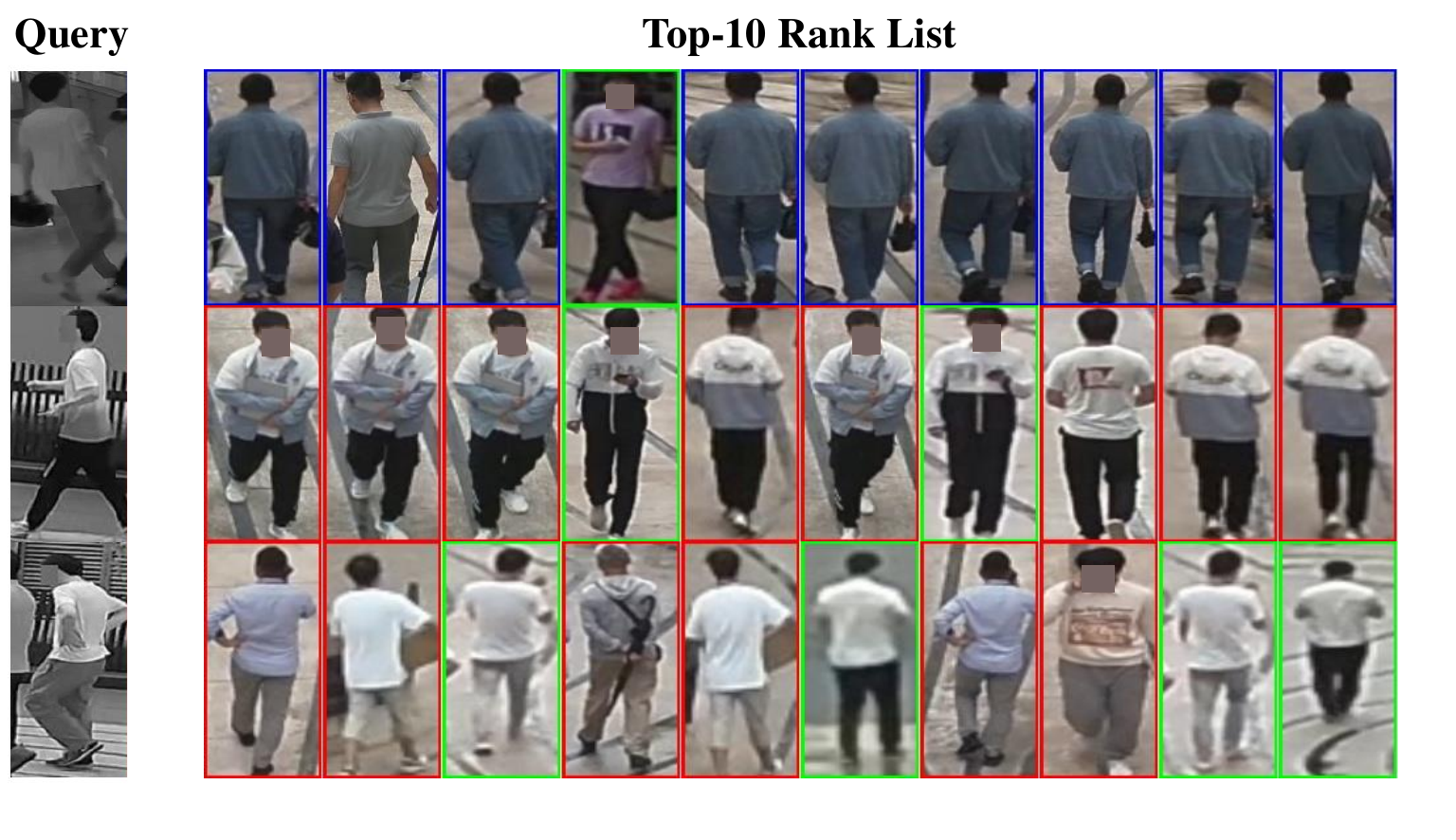}
    \caption{Visualization of OTLA-ReID retrieval results. Query images are from ground infrared cameras and gallery images are from UAV RGB cameras. Green, red and blue rectangles indicate correct, wrong retrieval results and distractors, respectively.}
    \label{fig:cctv_ir_uav_rgb}
\end{figure*}

\begin{figure*}[htbp]
    \centering
    \includegraphics[width=\linewidth]{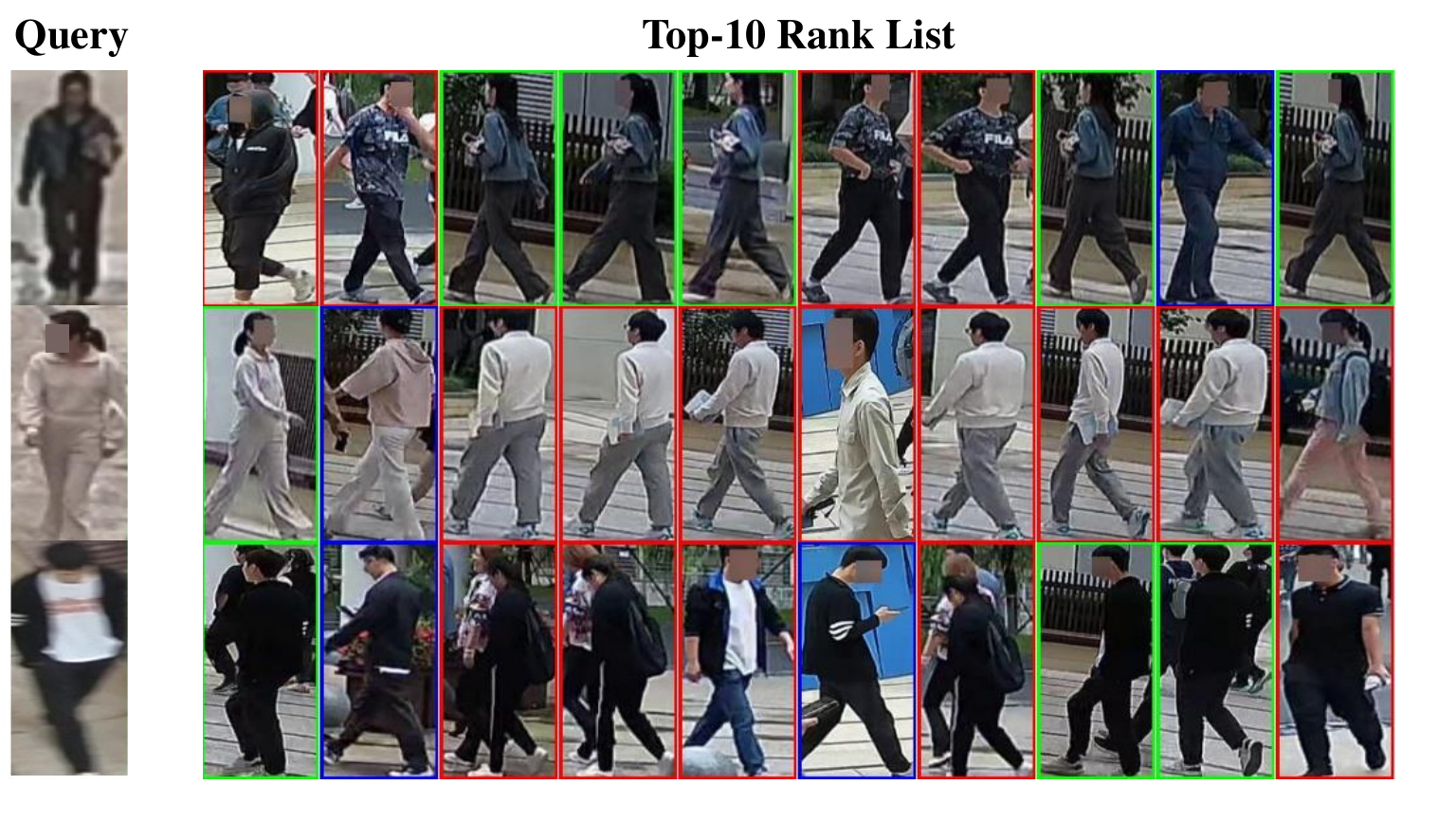}
    \caption{Visualization of OTLA-ReID retrieval results. Query images are from UAV RGB cameras and gallery images are from ground RGB cameras. Green, red and blue rectangles indicate correct, wrong retrieval results and distractors, respectively.}
    \label{fig:uav_rgb_cctv_rgb}
\end{figure*}

\begin{figure*}[htbp]
    \centering
    \includegraphics[width=\linewidth]{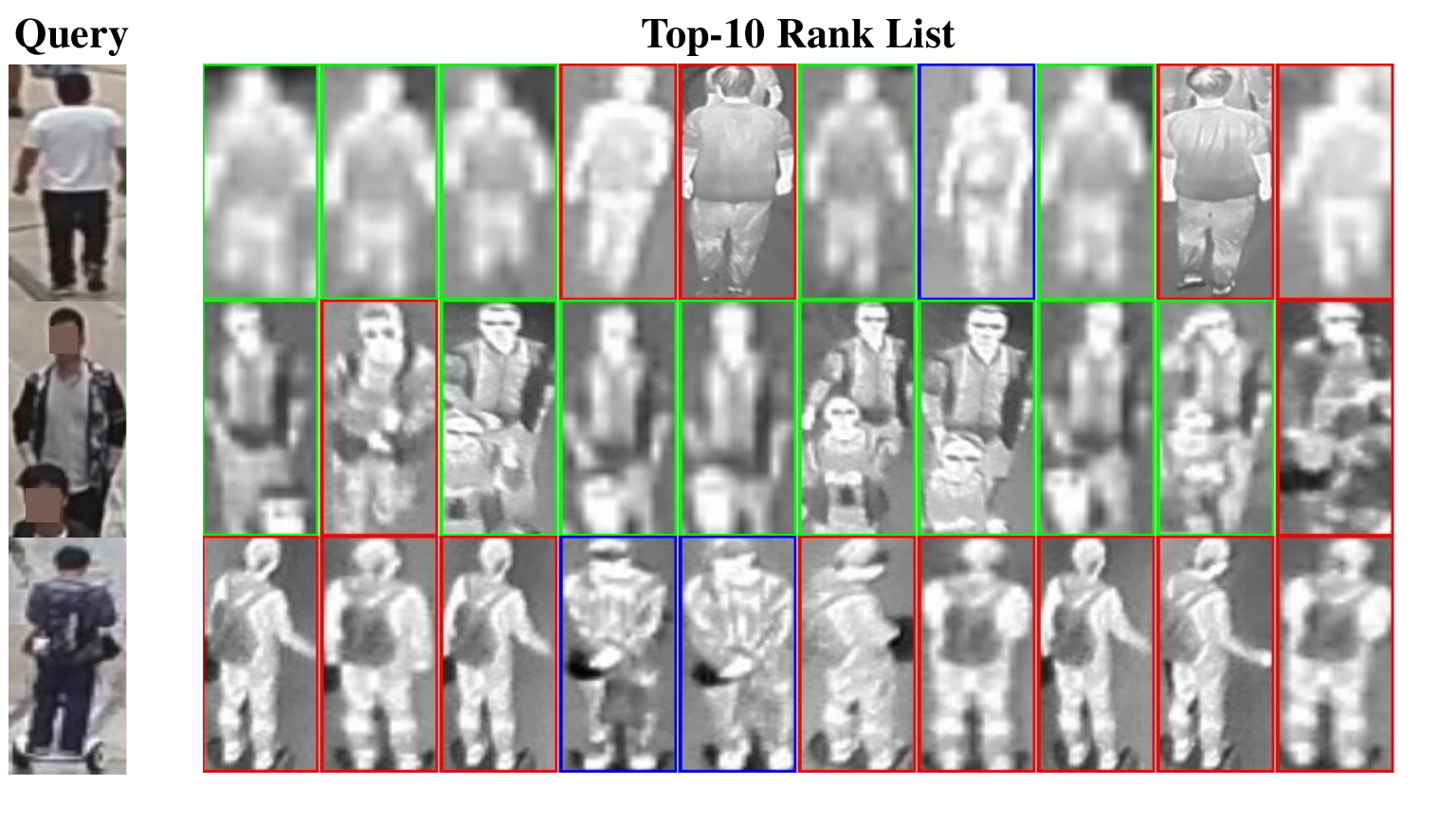}
    \caption{Visualization of OTLA-ReID retrieval results. Query images are from UAV RGB cameras and gallery images are from UAV thermal cameras. Green, red and blue rectangles indicate correct, wrong retrieval results and distractors, respectively.}
    \label{fig:uav_rgb_uav_ir}
\end{figure*}

In Fig.~\ref{fig:cctv_ir_uav_rgb}, ~\ref{fig:uav_rgb_cctv_rgb} and ~\ref{fig:uav_rgb_uav_ir} we visually show the top 10 retreival results for the tasks of  $\text{G}_{I} \rightarrow \text{U}_{R}$ (cross-modal \& cross-platform), $\text{U}_{R} \rightarrow \text{G}_{R}$( cross-platform only), and $\text{U}_{R} \rightarrow \text{U}_{T}$(cross-modal only) respectively. We designate OTLA-ReID[29] as the representative baseline method due to its superior overall performance.
The green rectangles indicate correctly retrieved results, the red ones indicate wrongly retrieved results and the blue ones represent the distractors. As shown, we can observe that our proposed MP-ReID poses a significant challenge to the existing algorithm, particularly after the inclusion of distractors. For instance, as depicted in  Fig.~\ref{fig:cctv_ir_uav_rgb}, we can see that OTLA-ReID exhibits high error rates, which is also heavily influenced by distractors. This arises from the inherent challenges of handing a cross-modal task compunded by various introduced by images originating from different platforms. From Fig.\ref{fig:uav_rgb_cctv_rgb}, it becomes evident that the unique perspective provided by the UAV present considerable obstacles for the existing baseline method , which are not explictly designed to accommodate such intricacies. In Fig.~\ref{fig:uav_rgb_uav_ir}, we present the retrieval results from two different modalities obtained by the UAV. Due to the inherent characteristics of the UAV, such as low resolution imaging and mobility, aligning between the RGB and thermal modalities poses additional difficulties. 
In summary, the aforementioned points highlight the necessity of our proposed MP-ReID, as it can provides strong support for the development of more robust algorithm capable of handling broader ranges of scenes and modalities.

\section{Social Impact}
The development and deployment of multi-modality multi-platform person ReID dataset carry significant social impact. Incorporating data from various modalities and platforms can promote the development of more accurate and robust personal identification technologies. It greatly aids in crowd analysis, urban planning, and traffic management, thereby advancing the development of smart cities. However, the dataset carries the risk of being targeted for attacks, thereby raising concerns about privacy breaches.

\end{document}